\documentclass{article}

\usepackage[T1]{fontenc}
\usepackage{iclr2027_conference,times}
\usepackage{amsmath,amssymb,amsthm}
\usepackage{booktabs}
\usepackage{graphicx}
\usepackage{float}
\usepackage{multirow}
\usepackage{microtype}
\usepackage{xcolor}
\usepackage{hyperref}
\usepackage{url}
\usepackage{enumitem}   
\usepackage{amsmath, amssymb}
\newcommand{\E}{\mathbb{E}}
\hypersetup{hidelinks}

\AddToHook{env/table/begin}{%
  \setlength{\abovecaptionskip}{0pt}%
  \setlength{\belowcaptionskip}{8pt}%
}

\usepackage{amsmath}     
\usepackage{amssymb}     
\usepackage{amsthm}      
\usepackage{booktabs}    

\makeatletter
\@ifundefined{theorem}{%
  \theoremstyle{plain}
  \newtheorem{theorem}{Theorem}[section]
}{}
\@ifundefined{proposition}{%
  \theoremstyle{plain}
  \@ifundefined{c@theorem}{\newtheorem{proposition}{Proposition}[section]}%
                          {\newtheorem{proposition}[theorem]{Proposition}}%
}{}
\@ifundefined{lemma}{%
  \theoremstyle{plain}
  \@ifundefined{c@theorem}{\newtheorem{lemma}{Lemma}[section]}%
                          {\newtheorem{lemma}[theorem]{Lemma}}%
}{}
\@ifundefined{corollary}{%
  \theoremstyle{plain}
  \@ifundefined{c@theorem}{\newtheorem{corollary}{Corollary}[section]}%
                          {\newtheorem{corollary}[theorem]{Corollary}}%
}{}
\@ifundefined{assumption}{%
  \theoremstyle{definition}
  \@ifundefined{c@theorem}{\newtheorem{assumption}{Assumption}[section]}%
                          {}%
}{}
\@ifundefined{definition}{%
  \theoremstyle{definition}
  \@ifundefined{c@theorem}{\newtheorem{definition}{Definition}[section]}%
                          {}%
}{}
\@ifundefined{remark}{%
  \theoremstyle{remark}
  \@ifundefined{c@theorem}{\newtheorem{remark}{Remark}[section]}%
                          {}%
}{}
\makeatother

\makeatletter
\newcommand{\safeop}[2]{%
  \@ifundefined{#1}{\expandafter\DeclareMathOperator\csname #1\endcsname{#2}}{}}
\newcommand{\safeopstar}[2]{%
  \@ifundefined{#1}{\expandafter\DeclareMathOperator\expandafter*\csname #1\endcsname{#2}}{}}
\makeatother

\safeop{Var}{Var}
\safeop{Cov}{Cov}
\safeop{Corr}{Corr}
\safeop{supp}{supp}
\safeopstar{argmax}{arg\,max}
\safeopstar{argmin}{arg\,min}
\safeopstar{esssup}{ess\,sup}
\safeopstar{essinf}{ess\,inf}

\providecommand{\E}{\mathbb{E}}

\providecommand{\eps}{\varepsilon}

\newcommand{\NA}{\textemdash}

\title{Whether or Where? Tail Calibration Signals Finite-Budget Reward
Hacking in Best-of-\texorpdfstring{$N$}{N} Selection}

\iclrfinalcopy

\author{
  Cheng Gui$^{1}$, Yihua Shao$^{2}$, Lei Li$^{3}$, Jingcai Guo$^{4}$, Maolin Wang$^{1}$ \\
  \textnormal{$^{1}$City University of Hong Kong} \quad
  \textnormal{$^{2}$University of Science and Technology Beijing} \\
  \textnormal{$^{3}$University of Washington} \quad
  \textnormal{$^{4}$Hong Kong Polytechnic University}
}

\begin{document}
\maketitle

\begin{abstract}
Best-of-$N$ (BoN) inference returns the candidate with the highest proxy
reward, so a larger $N$ tests the proxy's extreme tail more than its average
ranking quality. We show that the exact BoN value is a Beta-weighted integral
of conditional true reward over within-prompt score quantiles. It follows
that BoN depends on scores only through their ranks. It also follows that
BoN can decline only if the calibration curve in the tail turns downward,
and the budget where the decline appears depends on where that turn lies.
Under a bounded additive model, the error tail sets the limit: Gaussian
errors keep improving, Laplace errors reach a plateau, and heavy-tailed
errors return to the prior mean. When error variance is largest at low true
reward, even Gaussian noise can push BoN below random selection. Synthetic
phase maps agree with this taxonomy. On real candidate pools spanning
several verifiable domains, base policies, and learned reward models, tail
failures are rare and all occur in code generation. A held-out tail
calibration statistic reliably predicts where a cross-fitted budget cap
helps. It does about as well as rank accuracy, and the cap is competitive
with a Soft-BoN baseline. Tail calibration therefore does not replace
ranking metrics. It is an audit that is unchanged by any monotone rescaling
of the reward scores and shows which budgets are at risk.
\end{abstract}

\section{Introduction}

Inference-time alignment is the pragmatic answer to a stubborn problem: retraining a policy is expensive, but compute at serving time is cheap \cite{stiennon2020learning, nakano2021webgpt}. The default recipe is best-of-$N$ (BoN) selection: sample candidates from a frozen policy, score them with a reward model, return the winner. No gradients, embarrassingly parallel, and quality that improves as the budget grows. It is the closest thing the field has to a free lunch.

The lunch has a catch. {BoN does not sample the reward model, it searches for its maximum.} Push $N$ high enough and the proxy score keeps climbing while verifiable correctness turns over and falls. We call this \emph{inference-time reward hacking}: a regression caused entirely by the selection budget, with not a single weight updated. It arrives by spending more compute and is fixed by spending less, which leaves a practitioner two questions. {\textbf{1) Diagnosis:}} is this reward model exposed at all? {\textbf{2) Budgeting:}} if so, at what $N$?

Neither can be answered from the score distribution as a whole, because BoN never looks at it as a whole. Mapping scores to within-prompt quantiles $u=F_S(S)$, the quantile BoN returns has density $p_N(u)=Nu^{N-1}$, which collapses onto a vanishing neighborhood of $u=1$; at $N=64$, half the selection mass sits above the $99$th percentile. \textbf{A defect confined to the top percentile can dictate an entire deployment while shifting any average by an amount no benchmark would flag.} Yet reward models are graded by preference accuracy, calibration, or mean margin \cite{lambert2024rewardbench, Frick2024}, all averaged over the region selection never visits, and mitigated by KL penalties, soft selection, or ensembling \cite{gao2023scaling, eisenstein2023helping, coste2023reward}, all applied at a uniform rate whether or not a task is at risk. {A team can pass every leaderboard check and still ship a budget past its own optimum.}

A useful diagnostic must therefore read only the top of the distribution, depend on ranks alone, separate exposure from scale, and end in a budget cap that survives held-out data. All but the last follow from one identity. With $R$ the true utility and $Q(u)=\E[R \mid F_S(S)=u]$ the reward response in quantile space,
\begin{equation}
V(N) \;=\; \int_0^1 Q(u)\, N u^{N-1}\, du ,
\label{eq:vn}
\end{equation}
so \textbf{every question about budget becomes a question about the upper end of a single curve}: whether $V$ peaks depends on whether $Q$ bends down, and where it peaks on how soon.

That shape has a clean cause. Under $S=R+\varepsilon$ with bounded $R$, the tail class of $\varepsilon$ decides everything. Gaussian errors drive $Q$ toward the top of the support of $R$ and Laplace tails settle into a fixed exponential tilt, but regularly varying tails eventually make an extreme score \emph{uninformative} about $R$, so $Q$ falls. Only the last case admits a finite optimal budget, and it does so as soon as some intermediate budget already helps. Correlation strength leaves the tail class untouched, moving only the quantile at which informativeness runs out. \textbf{The tail decides \emph{whether} hacking is possible, correlation decides \emph{where} it appears.}

This tells us what to measure: a \emph{tail excess} statistic confined to the upper rank region,
\begin{equation}
\mathcal{T}_\tau \;=\; \frac{1}{1-\tau}\int_\tau^1 \big(Q(u) - \E[R]\big)\, du ,
\label{eq:tail}
\end{equation}
whose sign flags exposure while ignoring the bulk, recovered by cross-fitting so the same data cannot both propose and confirm a defect. One half of a task then estimates $\mathcal{T}_\tau$ and picks a budget cap; the other half checks what that cap buys. {The output is not a damping coefficient applied everywhere, but a named task, a named model, and a number.}

We validate by releasing our own assumptions one layer at a time, from synthetic settings that match the analysis, through variants that strain and finally break its prior and a study perturbing only the reward model's input, to public correctness data spanning several domains and learned reward models. {Tail failure turns out to be sharply localized rather than ambient}, and {a reading from the top percentile converts into a budget cap that still pays off on data never used to choose it}.

\begin{itemize}[leftmargin=*]
\item We formalize a blind spot in reward-model evaluation, namely that BoN concentrates on a vanishing top quantile while standard metrics report bulk behavior, and show that the resulting risk factorizes into a tail class governing whether a finite optimal budget exists and a correlation scale governing where.
\item We turn this into a practical audit: a monotone-invariant tail excess statistic, a cross-fitted reward response curve, and an estimate-then-validate procedure returning a budget cap. Unlike uniform mitigations such as rank-Soft-BoN, it names the task and model at risk and confirms the decision out of sample.
\item We validate at three levels of externality, from assumption-matched synthetics with deliberately violated variants, through an input-controlled study on a frozen candidate pool, to public reward-model data independent of our generator and model choices.
\end{itemize}
\section{Related Work}

\paragraph{Reward-model overoptimization.}
Proxy optimization can improve predicted reward while degrading human or
oracle utility \cite{gao2023scaling}.  BoN policies have also been analyzed through
distribution shift, KL control, coverage, and win-rate bounds
\cite{Beirami2024,Huang2025}.  Reward-model ensembles can mitigate some
overoptimization regimes \cite{Coste2023}.  We study a narrower object: the
joint calibration of proxy-score rank and true utility inside a fixed
candidate pool.

\paragraph{Inference-time reward hacking and tail-aware selection.}
Kwa et al.~\cite{Kwa2024} analyze catastrophic Goodhart effects under
heavy-tailed misspecification.  Khalaf et al.~\cite{Khalaf2025} derive the
probability-integral-transform and Beta order-statistic scaffold used here,
study single-peaked inference-time hacking, and propose Best-of-Poisson and
HedgeTune.  We do not claim that scaffold as a new theorem: our contribution is
to turn it into a score-transform-invariant joint-tail audit, separate tail
regime from finite-budget localization, and validate a held-out budget rule.
Huang et al.~\cite{Huang2025} propose InferenceTimePessimism (ITP), which
trades optimistic selection for a scaling-monotone pessimistic rule.  Most
recently, Hsu et al.~\cite{Hsu2026} propose Best-of-Tails, using a Hill
estimate of normalized reward-tail shape to choose a Tsallis-divergence
interpolation between optimistic and pessimistic selection.  Our contribution
is complementary: $E(u)$ estimates the \emph{joint} relation between true
utility and score rank, is invariant to strictly increasing score
transformations, and is evaluated as an audit-set budget trigger.  It requires
labeled outcomes, whereas a score-tail method can operate without them.

\paragraph{Reward-model evaluation.}
RewardBench, RM-Bench, and RMB expose capability, style, and bias failures
\cite{Lambert2024,Liu2024,Zhou2024}.  RewardBench 2 explicitly connects
multi-skill benchmark accuracy to downstream BoN and RLHF performance
\cite{Malik2025}.  PPE asks which offline reward-model metrics predict
post-RLHF outcomes and releases correctness data over MATH, GPQA, MMLU-Pro,
MBPP+, and IFEval \cite{Frick2024}; we reuse those public fixed candidate
pools for external validation.  At process level, PRM-BiasBench applies
semantics-preserving and semantics-altering perturbations to reasoning traces
and finds that fluent but corrupted steps can retain high reward
\cite{Bamba2025}.  Our masked and shuffled controls apply the same general
principle of controlled input perturbation to outcome-level reward models; we
do not evaluate intermediate-step PRMs.

\section{Problem Setup and Tail Representation}
\label{sec:theory}

We separate three questions usually conflated in discussions of
best-of-$N$ (BoN): \emph{whether} its value can turn down as $N$ grows,
\emph{where}, and \emph{how fast}.  Recommendations are collected in
Section~\ref{sec:prescribe}; proofs are in Appendix~\ref{app:proofs}.

\subsection{Score space, quantile space, and true utility}

For a prompt $x$, let $Y_1,\ldots,Y_N$ be conditionally i.i.d.\ responses
with true utility $R_i=R(x,Y_i)$ and proxy score $S_i=S(x,Y_i)$
($R_i=C_i\in\{0,1\}$ in verifiable experiments, real valued in the
synthetic study).  BoN selects
\begin{equation}
\label{eq:bon-selection}
 I_N=\argmax_{1\le i\le N}S_i,
 \qquad V_N=\E[R_{I_N}],
\end{equation}
with uniform random tie breaking.

For a continuous score CDF $F_S$, define
\begin{equation}
\label{eq:calibration-definitions}
 G(s)=\E[R\mid S=s],\qquad
 U=F_S(S),\qquad
 Q(u)=\E[R\mid U=u]=G(F_S^{-1}(u)),
\end{equation}
with $\mu=\E[R]$ and excess utility $E(u)=Q(u)-\mu$; $B_{a,b}$ denotes a
$\mathrm{Beta}(a,b)$ variable.  Empirically, $Q$ is estimated from
within-prompt ranks (tied blocks averaged) and averaged over prompts; all
statements below hold for this aggregate curve.

\begin{proposition}[Two identities]
\label{prop:quantile}
Let $S$ be continuous and $Q$ bounded.
\begin{enumerate}
\item[(i)] \textup{(BoN quantile identity)}
\begin{equation}
\label{eq:bon-quantile}
 V_N=\int_0^1 Q(u)\,Nu^{N-1}\,\mathrm{d}u=\E\big[Q(B_{N,1})\big],
 \qquad
 V_N-\mu=\int_0^1 E(u)\,Nu^{N-1}\,\mathrm{d}u .
\end{equation}
\item[(ii)] \textup{(Marginal-value identity)}
If $Q$ also has bounded variation, then for every $N\ge1$
\begin{equation}
\label{eq:marginal-value}
 \Delta V_N:=V_{N+1}-V_N=\int_0^1 u^{N}(1-u)\,\mathrm{d}Q(u),
\end{equation}
with $\mathrm{d}Q$ the Lebesgue--Stieltjes measure of $Q$; for absolutely
continuous $Q$,
$\Delta V_N=\E\big[Q'(B_{N+1,2})\big]\big/\big((N+1)(N+2)\big)$.
\end{enumerate}
\end{proposition}

Equation~\eqref{eq:bon-quantile} is standard \citep[cf.][]{Khalaf2025}: BoN sees the reward
model only through $Q$, so one pool gives $V_N$ at every budget up to its
size, invariant to strictly increasing recalibration.  By~\eqref{eq:marginal-value}, the
$(N{+}1)$-st sample earns exactly the increments of $Q$ it jumps over,
essentially near quantile $1-1/N$ (Appendix~\ref{app:proof-quantile}).

\subsection{Calibration shape constrains BoN}

For a signed measure $\kappa$ on $(0,1)$, $S^-(\kappa)$ is the smallest
$k$ such that $(0,1)$ splits into $k+1$ consecutive intervals on which
$\kappa$ alternates in sign (for piecewise monotone $Q$, $S^-(\mathrm{d}Q)$
counts turning points); for sequences, $S^-$ counts sign changes ignoring
zeros.

\begin{theorem}[Shape transfer]
\label{thm:shape}
Let $S$ be continuous and $Q$ of bounded variation.  Then
\begin{equation}
\label{eq:shape-transfer}
 S^-\big((\Delta V_N)_{N\ge1}\big)\;\le\;S^-(\mathrm{d}Q),
\end{equation}
with identical sign order under equality.  Consequently:
\begin{enumerate}
\item[(a)] if $Q$ is nondecreasing, so is $V_N$;
\item[(b)] if $Q$ is single-peaked (nondecreasing on $[0,u_\star]$,
nonincreasing on $[u_\star,1]$), so is $N\mapsto V_N$, with a finite peak
$N_\star$ if $Q$ strictly decreases on some $(u_1,1)$;
\item[(c)] $N\mapsto V_N$ has no more local extrema than $Q$.
\end{enumerate}
\end{theorem}

Under $t=-\log u$, $\Delta V_N$ is a Laplace transform of $\mathrm{d}Q$, so
by Descartes' rule of signs BoN can remove turning points of $Q$ but never
create them (Appendix~\ref{app:proof-shape}).  For single-peaked $Q$,
stopping at the first $\Delta V_N\le0$ is globally optimal.

The same shape constraint holds for finite pools: every turn in the
without-replacement BoN curve requires a turn in the tied-rank profile
(Proposition~\ref{prop:finitepool}, Appendix~\ref{app:proof-finite}).

If the top of $Q$ returns to the prompt mean after some finite budget
beats random, BoN has a finite maximizer (Corollary~\ref{prop:return},
Appendix~\ref{app:proof-return}).

\subsection{Additive noise and the prior tilt}
\label{sec:additive}

Consider $S=R+\eps$, with $R\in[r_{\min},r_{\max}]$ independent of $\eps$,
whose positive density $p_\eps$ has locally absolutely continuous logarithm
(e.g.\ $C^1$, or Laplace), and \emph{hazard slope}
$\psi_\eps(t)=-\tfrac{\mathrm{d}}{\mathrm{d}t}\log p_\eps(t)$ (a.e.).
The \emph{exponential tilt} of the prior,
\begin{equation}
\label{eq:exponential-tilt}
 m(\lambda)=\frac{\E[R\,e^{\lambda R}]}{\E[e^{\lambda R}]}
 =\frac{\mathrm{d}}{\mathrm{d}\lambda}\log\E\big[e^{\lambda R}\big],
 \qquad \lambda\in[0,\infty],
\end{equation}
is nondecreasing with $m(0)=\mu$, $m(\infty)=r_{\max}$.  The posterior at score $s$ is the prior tilted by the average
hazard slope over $[s-r,s]$; the posterior formula and binary specialization
are in Appendix~\ref{app:proof-cov} (Lemma~\ref{lem:tilt}).

Log-concave additive errors make $G$, $Q$, and $V_N$ nondecreasing for
every prior.  This follows from the covariance identity~\eqref{eq:covariance-identity} in Appendix~\ref{app:proof-cov}; for Gaussian
error, $G'(s)=\operatorname{Var}(R\mid S=s)/\sigma^2$.

Under additive noise, BoN can thus turn down only if $\psi_\eps$ decreases
somewhere, i.e.\ under heavy tails; log-concavity is also necessary for
monotonicity under every two-point prior (Appendix~\ref{app:proof-cov}).

\begin{theorem}[Hazard trichotomy]
\label{thm:hazard}
Assume $\lambda_\star:=\lim_{t\to\infty}\psi_\eps(t)\in[0,\infty]$ exists
and $\Pr(R>r_{\max}-\delta)>0$ for every $\delta>0$.  Then
\begin{equation}
\label{eq:hazard-limit}
 \lim_{s\to\infty}G(s)=\lim_{u\uparrow1}Q(u)=m(\lambda_\star):
\end{equation}
an extreme score reveals the prior tilted by the asymptotic hazard slope.
Explicitly,
\begin{itemize}
\item \emph{diverging influence}, $\lambda_\star=\infty$ (Gaussian,
generalised Gaussian $e^{-|t|^\beta}$, $\beta>1$): $Q(u)\to r_{\max}$;
\item \emph{constant influence}, $\lambda_\star\in(0,\infty)$ (Laplace with
scale $b$, $\lambda_\star=1/b$; logistic; $t^{k}e^{-\lambda_\star t}$;
$\beta=1$): $Q(u)\to m(\lambda_\star)\in(\mu,r_{\max})$, a plateau;
\item \emph{vanishing influence}, $\lambda_\star=0$ (Student-$t$, Pareto,
log-normal, $\beta<1$, and mixtures with such a component,
Proposition~\ref{prop:contam}): $Q(u)\to\mu$, so
Corollary~\ref{prop:return} applies once a finite budget beats random.
\end{itemize}
\end{theorem}

$\lambda_\star$ is the log-likelihood price of the extra error needed to
explain an extreme score without a high reward
(Appendix~\ref{app:proof-hazard}).  As it is also the limiting slope of
$-\log\bar F_\eps$, a log-survival plot of incorrect responses' scores
identifies the regime, which bulk metrics such as AUC cannot.

Rates of approach are given in Proposition~\ref{prop:rates}
(Appendix~\ref{app:proof-rates}).

\begin{table}[t]
\caption{Tail regimes for bounded rewards.  The first three rows assume
independent additive error (Theorem~\ref{thm:hazard}); Gaussian and Laplace
monotonicity follows from Proposition~\ref{prop:cov}.  The last row assumes
a unique scale maximizer (Proposition~\ref{prop:scale}).  Rates and further
families are given in Appendix~\ref{app:proof-rates}.}
\label{tab:trichotomy}
\centering
\small
\setlength{\tabcolsep}{4pt}
\begin{tabular}{llll}
\toprule
Error model & $\lambda_\star$ & Limiting utility & Budget behavior \\
\midrule
Gaussian & $\infty$ & $r_{\max}$ & Nondecreasing \\
Laplace (scale $b$) & $1/b$ & $m(1/b)$ & Nondecreasing; plateau \\
Student-$t_\nu$ & $0$ & $\mu$ & Returns to prior \\
Heteroskedastic Gaussian & \NA & $r_\sigma$ & Can end below $\mu$ \\
\bottomrule
\end{tabular}
\end{table}

\subsection{Heteroskedastic noise: scale beats shift}
\label{sec:hetero}

Neural reward-model errors may depend on $R$ and the prompt; the simplest
such departure already reverses the Gaussian conclusion.

\begin{proposition}[Scale beats shift]
\label{prop:scale}
Let $\eps\mid R=r\sim\mathcal N\big(0,\sigma^2(r)\big)$ with $\sigma\ge0$
continuous on $[r_{\min},r_{\max}]$, $\sigma(R)>0$ a.s., and a unique
maximiser $r_\sigma$ in the support of $R$.  Then
\begin{equation}
\label{eq:scale-limit}
 \lim_{s\to\infty}G(s)=r_\sigma,
 \qquad\text{hence}\qquad
 \lim_{N\to\infty}V_N=r_\sigma ,
\end{equation}
which lies \emph{below random selection} whenever $r_\sigma<\mu$.
\end{proposition}

The scale enters the log-likelihood at order $s^2$, the reward only at
order $s$ (Appendix~\ref{app:proof-scale}).  E.g.\ $S=R(1+\eps)$ has
$\sigma(r)\propto|r|$ and drives BoN to $r_{\min}$ when
$|r_{\min}|>r_{\max}$; our conditional-variance and multiplicative stress
tests probe these violations.

The model-free binary-reward characterization through $\Lambda=p_1/p_0$
is given in Proposition~\ref{prop:binaryLR} (Appendix~\ref{app:proof-scale}).

\subsection{Tail location sets the relevant budget}
\label{sec:where}

Where $V_N$ turns is set by how far into the top quantiles the offending
tail sits.

\begin{lemma}[Invisible tails]
\label{lem:invisible}
Let $Q,Q'$ take values in $[r_{\min},r_{\max}]$ and agree on
$[0,1-\delta]$.  Then for every $N$,
\begin{equation}
\label{eq:invisible-tails}
 |V_N-V'_N|\le(r_{\max}-r_{\min})\big(1-(1-\delta)^N\big)
 \le(r_{\max}-r_{\min})\,N\delta .
\end{equation}
\end{lemma}

Equation~\eqref{eq:invisible-tails} shows that budget $N$ reads $Q$ essentially on the top $1/N$ band, so the top
$\delta$ of $Q$ is invisible until $N\sim1/\delta$
(Appendix~\ref{app:proof-where}).

A heavy-tailed contaminating component can therefore preserve the observed
curve yet force an eventual return to the prior.  No finite pool certifies
monotonicity beyond its size (Proposition~\ref{prop:contam} and
Corollary~\ref{cor:nocert}, Appendix~\ref{app:proof-where}).

\subsection{Prescriptions}
\label{sec:prescribe}

Before choosing $N$, identify $\lambda_\star$ from the log-survival of
low-reward scores; when $Q$ is single-peaked, stop at the first budget
where the estimated $\Delta V_N$ (Proposition~\ref{prop:finitepool}) is
nonpositive.  Never extrapolate a rising curve beyond the pool size
(Corollary~\ref{cor:nocert}): in the plateau and heavy-tailed regimes, cap
$N$ near the empirical saturation point or peak, and pursue the Gaussian
regime's $\sqrt{\log N}$ gains only when samples are cheap.

\section{Experiments}
\label{sec:experiments}

We test three implications of Section~\ref{sec:theory}: error tails determine
limiting utility, calibration localizes finite-budget failures, and a
held-out tail audit can guide a budget cap.  Detailed settings, estimators,
and uncertainty procedures are in Appendix~\ref{app:protocol}.

\subsection{Experimental setup}
\label{sec:setup}
Synthetic pools use a bounded Uniform prior, Gaussian, Laplace, and
Student-$t$ errors, six correlations, and budgets up to $512$; prior and
multiplicative stress tests give 162 cells, with 12 additional
heteroskedastic cells.  Each cell uses 20,000 candidate sets.
For controlled tests, Qwen2.5-7B-Instruct \citep{Qwen2024} generates 256
answers per problem on MATH-500 and GSM8K \citep{Hendrycks2021,Cobbe2021};
four reward models score identical pools under QA and answer-only inputs,
with masked and shuffled MATH controls for two models.
PPE \citep{Frick2024} supplies 32 responses per prompt from four base
policies on five tasks, giving 65 learned task--RM cells and five verbosity
stress cells.  We compute exact without-replacement BoN curves with tied
blocks averaged.  A downturn requires a positive simultaneous 95\% interval
for the selected interior peak-to-endpoint drop.  In PPE, negative top-four
tail excess on a calibration fold triggers a calibration-selected budget
cap; the opposite fold evaluates it.  Policy tests use BH correction over
all 70 cells; diagnostic intervals bootstrap cells, preserving both folds.

\subsection{Synthetic tests of tail regime and budget scale}
\label{sec:synthetic}

\begin{figure}[t]
\centering
\includegraphics[width=.99\linewidth]{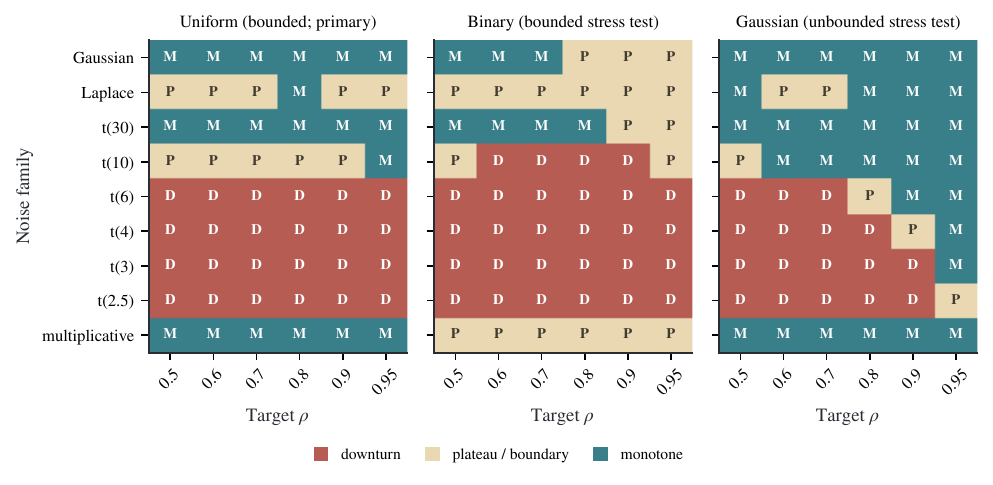}
\caption{\textbf{Synthetic phase maps.} D denotes an interior peak whose
peak-to-$N=512$ simultaneous max-$t$ interval is above zero; P denotes an
inconclusive interior or saturated boundary; M denotes a maximum at 512.  The
bounded Uniform map is primary.  Prior shape changes finite-window boundaries
but preserves the broad separation between light and heavy additive tails.}
\label{fig:phase}
\end{figure}

Figure~\ref{fig:phase} shows all 162 cells.  The bounded Uniform primary
prior satisfies the assumptions of Theorem~\ref{thm:hazard}, so every
additive error row has a predicted regime, and each is borne out.

\emph{Light-tailed additive errors.}  Gaussian cells are monotone
over $N\le512$, and by Proposition~\ref{prop:cov} this holds at every budget.
Laplace never produces a downturn: five cells are inconclusive plateaus
and one a boundary maximum.  Since Proposition~\ref{prop:cov} rules out a
true interior peak for Laplace error, its inconclusive interior peaks are
Monte Carlo fluctuations on the plateau $m(\lambda_\star)$ of
Theorem~\ref{thm:hazard}, which Proposition~\ref{prop:rates}(ii) says is
reached geometrically.

\emph{Heavy-tailed additive errors.}  The $t_{2.5}$, $t_3$, $t_4$, and $t_6$
rows have supported downturns at all six correlations, as the
vanishing-influence regime and Corollary~\ref{prop:return} require.
$t_{10}$ lies at the finite-window transition and $t_{30}$ is monotone.
Both belong to the same regime, but their return has rate $N^{-1/\nu}$
(Proposition~\ref{prop:rates}(i)): at $N=512$ this factor is still
$\approx0.81$ for $\nu=30$, against $\approx0.08$ for $\nu=2.5$, so the turn
lies beyond the window (Corollary~\ref{cor:nocert}).  Overall, 24/54
primary cells are downturns, 20 are monotone, and 10 are plateau/boundary.

Increasing correlation shifts and attenuates the heavy-tailed downturn
without changing its regime (Lemma~\ref{lem:invisible},
Theorem~\ref{thm:hazard}); detailed peak locations and intervals are in
Appendix~\ref{app:synthetic-location}.

Prior and multiplicative stress tests are reported in
Appendix~\ref{app:stress-results}; they clarify why the bounded-prior and
unique-scale-maximizer assumptions matter.

\emph{Heteroskedastic scale.}  Proposition~\ref{prop:scale} predicts
$V_N\to\sqrt3$ for $\beta=+1$ and $V_N\to-\sqrt3<\mu=0$ for $\beta=-1$.
All six increasing-scale cells peak at $N=512$, whereas four decreasing-scale
cells have supported downturns.  The largest drop, $1.939$, already takes
BoN below random selection; rank accuracy still reaches $0.828$ in a
failing cell.  Full contrasts are in Appendix~\ref{app:hetero-results}.

\subsection{The controlled MATH failure and format controls}
\label{sec:controlled-results}

Across the original 16 task--RM--format conditions, no QA curve has a
statistically supported downturn.  The clear original failure is Tulu on
MATH-500 answer-only: its peak moves from $N=64$ under QA to $N=4$, followed
by a $0.0597$ correctness drop at $N=256$ with simultaneous 95\% interval
$[0.0316,0.0878]$.  As Theorem~\ref{thm:shape} requires, the downturn is matched by a turn in the
calibration curve: the top-1\% excess changes sign from $E_1=+.054$ under QA
to $E_1=-.031$ (Table~\ref{tab:format-controls}), so the top of $Q$ falls
below the prompt mean.  GSM8K shows no supported downturn in either original
format, so information removal is not by itself sufficient; as for every
monotone curve here, this concerns only the window $N\le256$
(Corollary~\ref{cor:nocert}).

Section~\ref{sec:theory} identifies the change in $Q$ but not its cause, and
the format controls address the cause.  The failure does not survive a
role-preserving missing-question control: Tulu masked-question peaks at
$N=96$ with a nonsignificant $0.0082$ drop $[-0.0069,0.0232]$ and positive
top-1\% excess.  A shuffled question reduces rank accuracy and moves the
empirical peak to 16, but its $0.0098$ drop $[-0.0143,0.0339]$ is also
unsupported.  Thus the answer-only result cannot be attributed to missing
question semantics alone, and the chat role/template change is an important
confound in that condition.  Skywork is monotone when masked.  When
shuffled, its top-1\% excess falls to $+.005$, so the top of $Q$ has nearly
returned to the prompt mean, the situation of Corollary~\ref{prop:return};
the resulting peak at 8 is unsupported, providing a second model-level
negative control.

\begin{figure}[t]
\centering
\includegraphics[width=.98\linewidth]{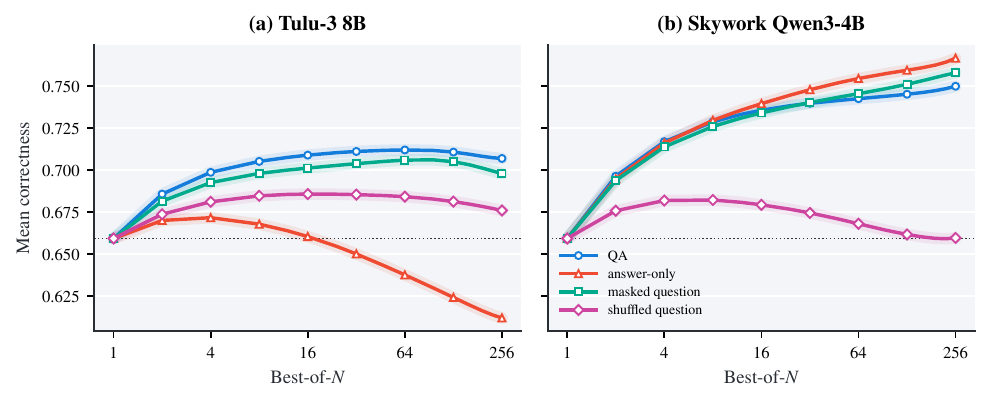}
\caption{\textbf{MATH-500 input-format controls on a fixed candidate pool.}
Each panel holds the reward model and all 500$\times$256 answers fixed while
varying only the score input: QA (blue circles), answer-only (orange-red triangles),
masked question (teal squares), or deterministically shuffled question
(magenta diamonds).  The thin
horizontal dotted line is base correctness.}
\label{fig:curves}
\end{figure}

\begin{table}[t]
\caption{MATH-500 input-format controls on the same 500$\times$256 candidate
pool.  $E_1$ is top-1\% tail excess; drop intervals are simultaneous within
each condition.  \NA\ indicates that the empirical maximum is at
$N=256$, so there is no selected interior peak contrast or corresponding
interval.}
\label{tab:format-controls}
\centering
\small
\setlength{\tabcolsep}{3.5pt}
\begin{tabular}{llrrrr}
\toprule
RM & Input & Rank acc. & $E_1$ & Peak $N$ & Drop [95\% CI] \\
\midrule
Tulu-3 8B & QA & .741 & +.054 & 64 & .0051 [-.0118,.0220] \\
 & answer-only & .581 & -.031 & 4 & .0597 [ .0316,.0878] \\
 & masked question & .698 & +.049 & 96 & .0082 [-.0069,.0232] \\
 & shuffled question & .613 & +.023 & 16 & .0098 [-.0143,.0339] \\
\midrule
Skywork Qwen3-4B & QA & .812 & +.084 & 256 & \NA \\
 & answer-only & .799 & +.101 & 256 & \NA \\
 & masked question & .792 & +.089 & 256 & \NA \\
 & shuffled question & .639 & +.005 & 8 & .0226 [-.0028,.0480] \\
\bottomrule
\end{tabular}
\end{table}

Exact tied-rank budget estimates and smoothing sensitivity are reported
in Appendix~\ref{app:controlled-localization}.  The answer-only optimum
remains near $N=3$--$4$ across visual estimators; unsupported conditions are
less stable.

\subsection{External validation: failures concentrate in code}
\label{sec:ppe-results}

\begin{figure}[t]
\centering
\includegraphics[width=.99\linewidth]{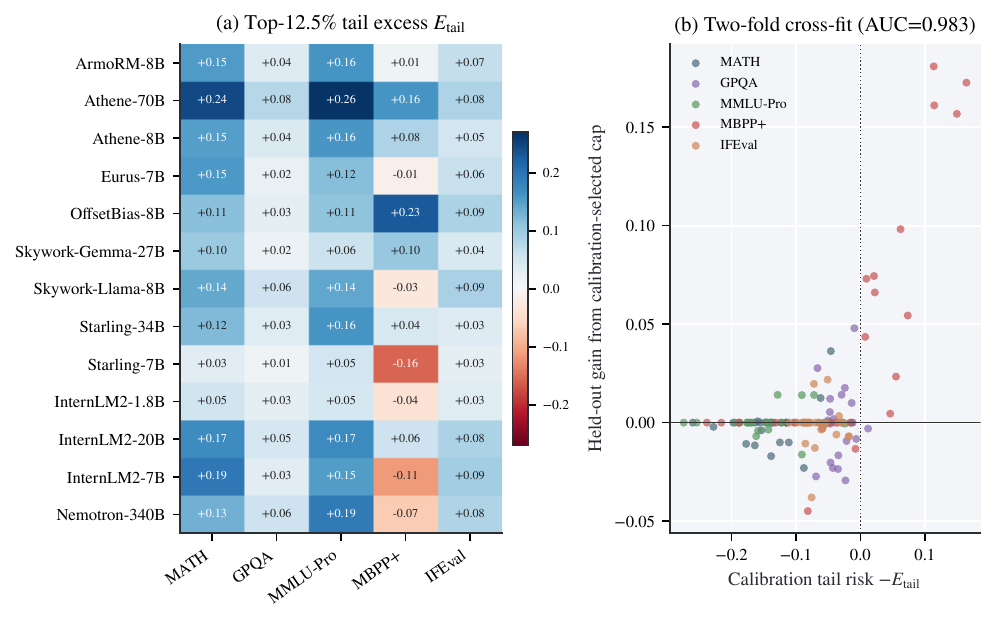}
\caption{\textbf{External PPE validation.} (a) Top-12.5\% tail excess for 13
learned reward models across five tasks.  Negative cells concentrate on
MBPP+.  (b) Calibration-fold tail risk versus held-out gain from the
calibration-selected cap, with two folds per learned cell.  Cell-clustered
bootstrap intervals are reported in Table~\ref{tab:diagnostics}.}
\label{fig:ppe}
\end{figure}

Figure~\ref{fig:ppe}a shows all 65 learned-model cells.  The six negative
top-12.5\% tail excesses all occur on MBPP+, and the other four tasks have
none.  By Proposition~\ref{prop:binaryLR}, a negative $E_{\rm tail}$ means
that incorrect programs outweigh correct ones in the top eighth of scores
($\Lambda<1$ there).  This public pool thus identifies a code-generation
failure domain, not a general upper-tail failure of current reward models.

The negative-tail cells show the two sign patterns that
Proposition~\ref{prop:finitepool} allows when the rank profile turns once.
If the profile first rises and then falls, the increments go $+$ then $-$,
giving an interior peak.  Eurus-RM-7B is the one learned cell satisfying our
strict interior-downturn definition: it peaks at $N=2$ and loses $0.059$
correctness by $N=32$, with simultaneous interval $[0.018,0.100]$.  If the
profile falls from the start, every increment is nonpositive and the peak is
the boundary $N=1$.  Three other negative-tail models are already harmful at
$N>1$ in this way: Starling-7B (drop $0.165$), InternLM2-7B ($0.171$), and
Nemotron-340B ($0.076$).  Keeping the two patterns apart prevents us from
hiding monotone harmful scaling behind the term ``downturn.''

Policy-stratified results and descriptive features of the bad code tail
are reported in Appendix~\ref{app:ppe-secondary}; they show that risk
depends on the policy--RM pairing and do not establish a causal shortcut.

\subsection{Held-out budget caps and diagnostic comparisons}
\label{sec:diagnostics}

Here we test the prescription of Section~\ref{sec:prescribe}: read the top
of $Q$, then cap $N$ if it has turned.  The two-fold analysis contains 130
learned-cell folds, nine of which have a calibration-selected cap with a
positive held-out interval: eight are MBPP+ and one is GPQA, for prevalence
$0.069$.  Tail risk predicts this rare event with ROC AUC $0.983$
[0.954, 0.998] and average precision (AP) $0.825$ [0.449, 0.973], versus the
prevalence baseline.

Rank accuracy is also strong (AUC/AP $0.972/0.747$); paired tail-minus-rank
intervals include zero for AUC and AP (Table~\ref{tab:diagnostics}).  The
broad AP intervals expose the rarity and reject any claim that bulk ranking
quality cannot predict the problem in practice; the connection to
non-identifiability is discussed in Appendix~\ref{app:diagnostic-interpretation}.

\begin{table}[t]
\caption{Diagnostic validation on 65 learned PPE cells.  Intervals resample
task--RM cells and keep both cross-fit folds together.  Differences are tail
risk minus the comparator.}
\label{tab:diagnostics}
\centering
\footnotesize
\setlength{\tabcolsep}{4pt}
\begin{tabular}{lrrr}
\toprule
Predictor & ROC AUC [95\% CI] & AP [95\% CI] & $\rho_s$ [95\% CI] \\
\midrule
$-E_{\mathrm{tail}}$ & .983 [.954,.998] & .825 [.449,.973] & .254 [.020,.459] \\
Low rank accuracy & .972 [.927,.995] & .747 [.337,.953] & .258 [.023,.453] \\
Score-tail spread & .355 [.133,.589] & .057 [.026,.117] & -.107 [-.292,.074] \\
\midrule
Tail $-$ rank & .012 [-.003,.042] & .079 [-.015,.215] & -.004 [-.043,.039] \\
Tail $-$ spread & .628 [.384,.858] & .769 [.382,.915] & .361 [.060,.656] \\
\bottomrule
\end{tabular}
\end{table}

What tail excess adds over rank accuracy is structural: it localizes the
failure to the budgets that read it (Lemma~\ref{lem:invisible}), it links
exactly to $V_N$ (Equation~\eqref{eq:bon-quantile}), and it is invariant
to monotone rescaling of scores.  A score-only tail spread performs poorly
(AUC $0.355$), as Proposition~\ref{prop:binaryLR} explains, because the marginal score shape does not
determine $\Lambda$.

Relative to BoN-32, the cap and rank-Soft-BoN gain $0.0085$ [0.0019, 0.0168]
and $0.0068$ [$-0.0003$, 0.0156] over all 65 cells (paired difference $0.0017$
[$-0.0003$, 0.0036]).  On six negative-tail cells they gain $0.0924$ and
$0.0916$: the hard rule is competitive, not materially superior.

Eurus, Starling-7B, InternLM2-7B, and Nemotron retain BH-corrected cap gains;
Skywork-Llama and InternLM2-1.8B have positive point gains only.  Unflagged
folds remain at $N=32$; consistent with Corollary~\ref{cor:nocert}, this
certifies nothing beyond the pool size (cell details:
Appendix~Table~\ref{tab:ppe-negative}).

Tail-width and tie-order checks preserve the main pattern
(Appendix~\ref{app:ppe-sensitivity}).  Missing endpoint normalization and
policy likelihoods prevent controlled Best-of-Tails and ITP comparisons;
rank-Soft-BoN is the feasible same-pool baseline.

\section{Conclusion}

We studied inference-time reward hacking through the tail calibration of
best-of-$N$ selection. The quantile representation explains how calibration
shape governs whether BoN can decline and how the location of tail defects
determines the relevant budget scale. For bounded rewards with independent
additive errors, the error tail distinguishes continued improvement,
saturation, and return to the prior mean. Synthetic and controlled
experiments support this account. In PPE, negative tail excess concentrates
in code generation, and cross-fitted budget caps yield held-out gains
competitive with rank-Soft-BoN. Tail calibration thus complements rank
accuracy with a direct connection between reward-model evaluation and
search budgets. With labeled candidate pools and independent validation,
this connection supports task--policy--reward-model-specific budget choices
grounded in true utility over the tested range.

\clearpage
\subsection*{AI use statement}

Generative AI tools assisted with literature
synthesis, manuscript drafting and editing, code organization, figure
preparation, and LaTeX formatting.  Experimental commands, retained
artifacts, numerical claims, references, and rendered output were checked
against repository or primary-source evidence.  No model-generated numerical
result was accepted without a corresponding analysis artifact.  The authors
are responsible for the final content, including text, claims, and artifacts
produced with generative AI assistance.

\subsection*{Ethics statement}

The work uses no personal data and does not target deployed systems.  Because
a tail diagnostic could help optimize against a weak reward model, we frame it
as a pre-deployment audit requiring independent ground-truth verification.

\subsection*{Reproducibility statement}

The assumptions and proof details are provided in
Appendix~\ref{app:proofs}; data processing, estimators, and experimental
settings are described in Appendix~\ref{app:protocol}, with sensitivity
analyses in Appendix~\ref{app:ppe-sensitivity}.  Complete controlled results and additional PPE results appear in
Appendices~\ref{app:controlled-results} and~\ref{app:ppe-results}.

\bibliographystyle{iclr2027_conference}
\bibliography{references}

\appendix
\setcounter{totalnumber}{4}

\section{Proofs for Section~\ref{sec:theory}}
\label{app:proofs}

The results of Section~\ref{sec:theory} rest on two mechanisms, and the
proofs are organised around them.
\begin{itemize}
\item \emph{Order statistics and total positivity}
(Appendices~\ref{app:proof-quantile}--\ref{app:proof-return}).  BoN reads
the calibration curve at the random quantile $U_{(N)}$, and one extra
sample reads its increments through the kernel $u^N(1-u)$.  Because this
kernel is totally positive in $(N,u)$, aggregation over $N$ cannot create
sign changes.
\item \emph{Explaining away at extreme scores}
(Appendices~\ref{app:proof-cov}--\ref{app:proof-scale}).  The posterior
over reward at a large score is a tilt of the prior whose rate is the
hazard slope of the error (additive noise) or, more generally, the growth
rate of the class-conditional log-likelihoods.
\end{itemize}
Appendix~\ref{app:proof-where} combines the two to separate whether a
failure occurs from where it becomes visible.

\paragraph{Conventions.}
(C1) \emph{Versions.}  $Q$ is a conditional expectation and is defined only
up to Lebesgue-null sets of $u$.  Every quantity we compute ($V_N$,
$\Delta V_N$) integrates $Q$ against a law with a density on $(0,1)$, so it
does not depend on the version.  When $Q$ has bounded variation we use its
right-continuous version on $(0,1)$; then $\mathrm{d}Q((a,b])=Q(b)-Q(a)$,
the limits $Q(0^+)$ and $Q(1^-)$ exist, and $|\mathrm{d}Q|$ has finite total
mass $\mathrm{TV}(Q)$.
(C2) \emph{Sign changes.}  For a signed measure $\kappa$ on $(0,1)$ with
$S^-(\kappa)=k<\infty$ we call the $k+1$ intervals of a minimal partition
the \emph{sign blocks} of $\kappa$; by minimality each block carries nonzero
mass and consecutive blocks carry opposite signs.
(C3) \emph{Total positivity} \citep{Karlin1968}.  A kernel $K(x,y)$ on
ordered sets $X\times Y$ is totally positive (TP) if every minor
$\det[K(x_i,y_j)]_{i,j\le m}$ with $x_1<\dots<x_m$, $y_1<\dots<y_m$ is
nonnegative, and strictly totally positive (STP) if every such minor is
positive.
(C4) \emph{Beta facts.}  $U_{(N)}\sim\mathrm{Beta}(N,1)$ and
$1-U_{(N)}\sim\mathrm{Beta}(1,N)$.  $B_{N+1,2}$ has mode $N/(N+1)$, mean
$1-2/(N+3)$ and standard deviation
$\sqrt{2(N+1)}\big/\big((N+3)\sqrt{N+4}\big)=\tfrac{\sqrt2}{N}(1+O(1/N))$.

\subsection{Proof of Proposition~\ref{prop:quantile}}
\label{app:proof-quantile}

\paragraph{Preliminaries.}
Because $S$ is continuous, $\Pr(S_i=S_j)=0$ for $i\ne j$, so ties have
probability zero and the tie-breaking rule plays no role.  The variables
$U_i=F_S(S_i)$ are i.i.d.\ Uniform$[0,1]$, and $S_i>S_j$ implies
$U_i\ge U_j$, with equality on a null event.  Hence, almost surely, $I_N$
is also the index of $U_{(N)}$.  Nothing beyond the ranking of the scores
is used below, which is the source of the invariance of every statement
under strictly increasing transformations of $S$.

\paragraph{Proof of Equation~\eqref{eq:bon-quantile}.}
By exchangeability of the pairs $(R_i,U_i)$,
\begin{equation}
\label{eq:winner-exchangeability}
 V_N=\sum_{i=1}^N\E\big[R_i\,\mathbf 1\{I_N=i\}\big]
    =N\,\E\big[R_1\,\mathbf 1\{U_1>U_j,\ 2\le j\le N\}\big].
\end{equation}
The variables $U_2,\ldots,U_N$ are independent of $(R_1,U_1)$, so
conditioning on $(R_1,U_1)$ gives
$\Pr(U_1>U_j\ \forall j\ge2\mid R_1,U_1)=U_1^{N-1}$ and
\begin{equation}
\label{eq:quantile-proof}
 V_N=N\,\E\big[R_1U_1^{N-1}\big]=N\,\E\big[\E[R_1\mid U_1]\,U_1^{N-1}\big]
    =\int_0^1Q(u)\,Nu^{N-1}\,\mathrm{d}u .
\end{equation}
The second form follows from $\int_0^1Nu^{N-1}\,\mathrm{d}u=1$.
Equation~\eqref{eq:quantile-proof} says that the winner's reward has the law of a reward read at
the random quantile $U_{(N)}\sim\mathrm{Beta}(N,1)$: the budget acts only by
moving this reading point towards $u=1$.

\paragraph{Proof of Equation~\eqref{eq:marginal-value}: the crossing argument.}
Realise the $(N{+}1)$-sample experiment by adding one fresh sample to the
$N$-sample experiment, so that $U_{(N+1)}=\max(U_{(N)},U_{N+1})$ with
$U_{N+1}$ independent of $U_{(N)}$.  By~\eqref{eq:bon-quantile} applied to $N$ and $N+1$,
\begin{equation}
\label{eq:marginal-crossing}
 \Delta V_N=\E\big[Q(U_{(N+1)})-Q(U_{(N)})\big]
 =\E\Big[\big(Q(U_{N+1})-Q(U_{(N)})\big)\,\mathbf 1\{U_{N+1}>U_{(N)}\}\Big].
\end{equation}
On the event $\{U_{N+1}>U_{(N)}\}$ the bracket equals
$\mathrm{d}Q\big((U_{(N)},U_{N+1}]\big)
=\int_{(0,1)}\mathbf 1\{U_{(N)}<u\le U_{N+1}\}\,\mathrm{d}Q(u)$.  Since
$|\mathrm{d}Q|$ is finite, Fubini's theorem applies and
\begin{equation}
\label{eq:marginal-kernel}
 \Delta V_N=\int_{(0,1)}\Pr\big(U_{(N)}<u\le U_{N+1}\big)\,\mathrm{d}Q(u)
 =\int_{(0,1)}u^N(1-u)\,\mathrm{d}Q(u).
\end{equation}
The proof explains the kernel.  A new sample changes the outcome only if
it dethrones the incumbent, and it then earns the increments of $Q$ over
the quantile interval it jumps across.  The kernel $u^N(1-u)$ is the
probability that level $u$ is jumped across: the incumbent is still below
$u$ (probability $u^N$, which is small unless $u$ is close to $1$) and the
newcomer clears $u$ (probability $1-u$, which vanishes at $u=1$).  The
product peaks at $u=N/(N+1)$, so the marginal sample can change the value
only through the part of $Q$ near the current record.

\paragraph{Second proof of Equation~\eqref{eq:marginal-value} and the limit of $V_N$.}
For $Q$ of bounded variation and the continuous function $g(u)=u^N$,
Lebesgue--Stieltjes integration by parts has no jump correction and gives
$\int_{(0,1)}Q\,\mathrm{d}g=Q(1^-)g(1)-Q(0^+)g(0)-\int_{(0,1)}g\,\mathrm{d}Q$,
that is,
\begin{equation}
\label{eq:ibp}
 V_N=Q(1^-)-\int_{(0,1)}u^N\,\mathrm{d}Q(u).
\end{equation}
Differencing~\eqref{eq:ibp} in $N$ gives~\eqref{eq:marginal-value} again.  Since $u^N\to0$ on
$(0,1)$ and $|\mathrm{d}Q|$ is finite, dominated convergence in
\eqref{eq:ibp} gives $V_N\to Q(1^-)$: the BoN limit is the top of the
calibration curve, and the remaining gap $\int u^N\,\mathrm{d}Q$ is a
Laplace-type transform of $\mathrm{d}Q$ concentrated near $u=1$.

\paragraph{Beta form and the local question.}
If $Q$ is absolutely continuous, $\mathrm{d}Q=Q'(u)\,\mathrm{d}u$ and
$\int_0^1u^N(1-u)Q'(u)\,\mathrm{d}u=\mathrm{B}(N+1,2)\,\E[Q'(B_{N+1,2})]$,
where $\mathrm{B}(N+1,2)=\Gamma(N+1)\Gamma(2)/\Gamma(N+3)=1/((N+1)(N+2))$.
By (C4), $B_{N+1,2}$ sits at $1-\Theta(1/N)$ with spread $\asymp1/N$, so
$(N+1)(N+2)\,\Delta V_N$ is an average of the slope $Q'$ over a window of
width $\asymp1/N$ around $1-1/N$; this is the ``local question'' of the
main text.

\paragraph{Prompt averaging.}
If $V_N(x)=\int_0^1Q_x(u)\,Nu^{N-1}\,\mathrm{d}u$ for each prompt $x$ and
the $Q_x$ are uniformly bounded, Fubini's theorem gives
$\E_x V_N(x)=\int_0^1\bar Q(u)\,Nu^{N-1}\,\mathrm{d}u$ with
$\bar Q=\E_xQ_x$.  If moreover $\E_x\mathrm{TV}(Q_x)<\infty$, then
$\mathrm{d}\bar Q=\E_x\,\mathrm{d}Q_x$ and~\eqref{eq:marginal-value} holds for the average.
\hfill$\square$

\subsection{Finite pools: proof of Proposition~\ref{prop:finitepool}}
\label{app:proof-finite}

\begin{proposition}[Finite pools]
\label{prop:finitepool}
Let $\bar r_{(1)},\ldots,\bar r_{(K)}$ be the prompt-averaged rewards at
ascending within-prompt score ranks (tied blocks averaged, pool size $K$),
and $\hat V_N$ the without-replacement BoN curve.  Then
\begin{equation}
\label{eq:finite-pool-increments}
 \hat V_{N+1}-\hat V_N=\sum_{j=1}^{K-1}k_N(j)\,\big(\bar r_{(j+1)}-\bar r_{(j)}\big),
 \qquad
 k_N(j)=\frac{\binom{j}{N}(K-j)}{\binom{K}{N}(K-N)}\ge0,
\end{equation}
where $k_N(j)$ is the probability that the best of the first $N$ draws has
rank at most $j$ and the next draw rank above $j$; moreover
$S^-\big((\hat V_{N+1}-\hat V_N)_{N<K}\big)\le
S^-\big((\bar r_{(j+1)}-\bar r_{(j)})_{j<K}\big)$.
\end{proposition}

Every turn in a reported BoN curve is thus witnessed by a turn in the rank
profile (Appendix~\ref{app:proof-finite}).

Fix a prompt with $K$ responses and let $r_{(1)},\ldots,r_{(K)}$ be their
rewards ordered by ascending score, with tied blocks averaged.  BoN at
budget $N\le K$ draws $N$ of the $K$ responses uniformly without
replacement and returns the reward of the highest-ranked draw.

\begin{lemma}[Finite-pool identities]
\label{lem:finite}
Let $d_j=r_{(j+1)}-r_{(j)}$ and $W_N(j)=\binom{j}{N}/\binom{K}{N}$ (with
$\binom{j}{N}=0$ for $j<N$).  Then
\begin{enumerate}
\item[(i)] $\hat V_N=\sum_{j=N}^{K}\binom{j-1}{N-1}\binom{K}{N}^{-1}r_{(j)}
 =r_{(K)}-\sum_{j=1}^{K-1}W_N(j)\,d_j$;
\item[(ii)] for $N\le K-1$,
 $\hat V_{N+1}-\hat V_N=\sum_{j=1}^{K-1}k_N(j)\,d_j$ with
 $k_N(j)=W_N(j)\,\frac{K-j}{K-N}\ge0$;
\item[(iii)] $S^-\big((\hat V_{N+1}-\hat V_N)_{N=1}^{K-1}\big)\le
 S^-\big((d_j)_{j=1}^{K-1}\big)$.
\end{enumerate}
All three statements continue to hold for prompt averages with a common
$K$, with $r_{(j)}$ replaced by $\bar r_{(j)}$.
\end{lemma}

\begin{proof}
(i) The top draw has rank $j$ exactly when it is $j$ and the other $N-1$
draws come from the $j-1$ responses below it, which has probability
$\binom{j-1}{N-1}/\binom{K}{N}$.  With uniform tie breaking the expected
selected reward is linear in the reward vector, and averaging over the
random order inside a tied block replaces each reward of the block by the
block average.  The second form is summation by parts, using
$W_N(j)-W_N(j-1)=\binom{j-1}{N-1}/\binom{K}{N}$, $W_N(K)=1$ and $W_N(0)=0$.

(ii) By (i), $\hat V_{N+1}-\hat V_N=\sum_j\big(W_N(j)-W_{N+1}(j)\big)d_j$.
Draw the $N+1$ responses one at a time.  $W_N(j)$ is the probability that
the first $N$ all have rank at most $j$, and $W_{N+1}(j)$ is the
probability that all $N+1$ do, so the difference is the probability that
the first $N$ have rank at most $j$ and the next one has rank above $j$.
Given the first event, the next draw is uniform over the $K-N$ remaining
responses, of which $K-j$ lie above $j$; hence
$W_N(j)-W_{N+1}(j)=W_N(j)(K-j)/(K-N)$.  This is the exact finite-pool
counterpart of the crossing kernel $u^N(1-u)$, to which it converges when
$j/K\to u$ with $N$ fixed.

(iii) As a matrix indexed by $(N,j)$,
$k_N(j)=a_N\,b_j\binom{j}{N}$ with
$a_N=1/\big(\binom{K}{N}(K-N)\big)>0$ and $b_j=K-j>0$ for $j\le K-1$.
Multiplying rows and columns by positive numbers does not change the sign
of any minor, so $k$ is TP whenever the Pascal kernel
$(N,j)\mapsto\binom{j}{N}$ is.  The Pascal matrix is totally nonnegative:
each minor counts families of non-intersecting lattice paths by the
Lindstr\"om--Gessel--Viennot lemma.  The variation-diminishing property of
TP kernels \citep[Ch.~5, Thm.~3.1]{Karlin1968} then gives (iii).  Prompt
averaging is linear with a common kernel, which proves the last claim.
\end{proof}

The lemma makes the diagnostic of Theorem~\ref{thm:shape} exact at every
sample size.  A turn in the reported BoN curve must be matched by a turn in
the reported rank profile, whatever the sampling noise, because the curve is
a TP transform of the profile's increments.  Noise can only add sign
changes to the profile; it cannot hide the witness of a turn.

\subsection{Proof of Theorem~\ref{thm:shape}}
\label{app:proof-shape}

We give a self-contained proof in three steps.  Step~2 is Laguerre's proof
of Descartes' rule of signs, transplanted to the kernel $u^\lambda(1-u)$; an
alternative is to note that $(N,u)\mapsto u^N(1-u)=(1-u)e^{N\log u}$ is STP,
because $e^{xy}$ is STP and $\log u$ is increasing, and to invoke the
variation-diminishing theorem \citep[Ch.~5, Thm.~3.1]{Karlin1968}.

Let $\mathcal M$ be the class of signed measures $\kappa$ on $(0,1)$ with
$\int u^\lambda(1-u)\,|\kappa|(\mathrm{d}u)<\infty$ for every $\lambda>0$,
and for $\kappa\in\mathcal M$ put
\begin{equation}
\label{eq:signed-transform}
 \Phi_\kappa(\lambda)=\int_{(0,1)}u^\lambda(1-u)\,\kappa(\mathrm{d}u),
 \qquad\lambda>0 .
\end{equation}
Every finite signed measure, in particular $\mathrm{d}Q$, lies in
$\mathcal M$, and by Equation~\eqref{eq:marginal-value}
$\Delta V_N=\Phi_{\mathrm{d}Q}(N)$.  The idea is to embed the sequence
$(\Delta V_N)$ in the real-analytic function $\Phi_{\mathrm{d}Q}$ and count
zeros of the function instead of sign changes of the sequence.

\paragraph{Step 1: sign changes of the sequence are zeros of the function.}
$\Phi_\kappa$ is continuous on $(0,\infty)$ by dominated convergence.  If
$\Phi_\kappa(N_1)$ and $\Phi_\kappa(N_2)$ have opposite signs, with
$N_1<N_2$ and $\Phi_\kappa(N)=0$ for the integers strictly between them,
then $\Phi_\kappa$ vanishes somewhere in the open interval $(N_1,N_2)$,
either at one of those integers or by the intermediate value theorem.
Distinct sign changes of the sequence give zeros in disjoint intervals, so
\begin{equation}
\label{eq:sequence-zero-bound}
 S^-\big((\Phi_\kappa(N))_{N\ge1}\big)\le
 \#\{\lambda>0:\Phi_\kappa(\lambda)=0\}.
\end{equation}

\paragraph{Step 2: the function has at most $S^-(\kappa)$ zeros.}
We show by induction on $k$ that if $\kappa\in\mathcal M$ is nonzero and
$S^-(\kappa)\le k$, then $\Phi_\kappa$ has at most $k$ distinct zeros on
$(0,\infty)$.  For $k=0$, $\kappa$ or $-\kappa$ is a nonzero nonnegative
measure and the integrand $u^\lambda(1-u)$ is strictly positive on $(0,1)$,
so $\Phi_\kappa$ never vanishes.  For the induction step let
$S^-(\kappa)=k\ge1$ and let $c\in(0,1)$ be the point that separates the
last two sign blocks.  Put
\begin{equation}
\label{eq:rescaled-transform}
 \Psi(\lambda)=c^{-\lambda}\Phi_\kappa(\lambda)
 =\int(u/c)^\lambda(1-u)\,\kappa(\mathrm{d}u),
\end{equation}
which has the same zeros as $\Phi_\kappa$.  For $\lambda$ in a compact
subinterval $[\lambda_0,\lambda_1]\subset(0,\infty)$,
\begin{equation}
\label{eq:transform-derivative-bound}
 \big|\partial_\lambda(u/c)^\lambda\big|=|\log(u/c)|\,(u/c)^\lambda\le C\,u^{\lambda_0/2}
 \qquad\text{on }(0,1),
\end{equation}
so differentiation under the integral is permitted and
\begin{equation}
\label{eq:transform-derivative}
 \Psi'(\lambda)=c^{-\lambda}\,\Phi_{\kappa'}(\lambda),
 \qquad \kappa'(\mathrm{d}u)=\log(u/c)\,\kappa(\mathrm{d}u).
\end{equation}
The same bound shows $\kappa'\in\mathcal M$.  The factor $\log(u/c)$ is
negative on $(0,c)$ and positive on $(c,1)$.  It therefore flips the sign
of every block to the left of $c$ and keeps the last block, so the last two
blocks now carry the same sign and merge, while all other sign changes
survive.  Hence $S^-(\kappa')\le k-1$.  Moreover $\kappa'\ne0$, because
$\kappa$, having $k\ge1$ sign changes, is not concentrated at the single
point $c$.  By the induction hypothesis $\Psi'$ has at most $k-1$ zeros.
If $\Phi_\kappa$, and hence $\Psi$, had $k+1$ distinct zeros, Rolle's
theorem would give $k$ distinct zeros of $\Psi'$, a contradiction.

Multiplying by $\log(u/c)$ is the familiar ``multiply by $e^{c\lambda}$ and
differentiate'' step of Descartes' rule, written in the variable
$t=-\log u$: each differentiation removes exactly one sign change, and
Rolle's theorem says it removes at most one zero.  Steps~1 and~2 give
$S^-\big((\Delta V_N)_{N\ge1}\big)\le S^-(\mathrm{d}Q)$, which is the Laplace-transform version of the same sign-change bound.

\paragraph{Step 3: the sign at large $N$, and equality.}
Let $S^-(\mathrm{d}Q)=k$, let the last sign block be $(c_k,1)$, and let
$\epsilon_{k+1}\in\{-1,+1\}$ be its sign.  Split
$\Phi_{\mathrm{d}Q}(\lambda)$ into the integral over $(0,c_k]$, bounded in
absolute value by $c_k^\lambda\,\mathrm{TV}(Q)$, and the integral over
$(c_k,1)$, which has sign $\epsilon_{k+1}$.  The last block has nonzero
mass, so there are $c_k<v<v'<1$ with $|\mathrm{d}Q|((v,v'))>0$, and the
second integral is at least $v^\lambda(1-v')\,|\mathrm{d}Q|((v,v'))$ in
absolute value.  Since $v>c_k$, the second integral dominates:
\begin{equation}
\label{eq:last-sign}
 \operatorname{sign}\Delta V_N=\epsilon_{k+1}\qquad\text{for all large }N .
\end{equation}
The tail of the BoN increments thus carries the sign of the top of the
calibration curve.  If $S^-(\Delta V)=k$ as well, the two sign sequences
have the same length and the same last element, and alternating sequences
are determined by these, so the patterns coincide.

\paragraph{Consequences.}
(a) If $Q$ is nondecreasing, $\mathrm{d}Q\ge0$ and $\Delta V_N\ge0$
directly from Equation~\eqref{eq:marginal-value}; no total positivity is
needed.
(c) The local extrema of $N\mapsto V_N$ are the sign changes of
$(\Delta V_N)$, and for piecewise monotone $Q$ the turning points of $Q$
are the sign changes of $\mathrm{d}Q$.

(b) The single-peaked case has a one-line proof that shows the mechanism
directly.  Let $\mathrm{d}Q\ge0$ on $(0,u_\star]$ and $\le0$ on
$(u_\star,1)$, with $u_\star>0$; the case $u_\star=0$ is trivial.  Dividing
by $u_\star^N$,
\begin{equation}
\label{eq:single-peak-decomposition}
 \frac{\Delta V_N}{u_\star^N}
 =\underbrace{\int_{(0,u_\star]}\Big(\frac{u}{u_\star}\Big)^N(1-u)\,\mathrm{d}Q(u)}_{A_N\ \ge0}
 \;-\;\underbrace{\int_{(u_\star,1)}\Big(\frac{u}{u_\star}\Big)^N(1-u)\,|\mathrm{d}Q|(\mathrm{d}u)}_{B_N\ \ge0}.
\end{equation}
Since $(u/u_\star)^N$ is nonincreasing in $N$ for $u\le u_\star$ and
nondecreasing for $u>u_\star$, $A_N$ is nonincreasing and $B_N$ is
nondecreasing, so $A_N-B_N$ is nonincreasing in $N$.  The set
$\{N:\Delta V_N\le0\}$ is therefore an up-set: once one more sample stops
helping, it never helps again, $V_N$ is single-peaked, and
$N_\star=\min\{N:\Delta V_N\le0\}$ is a global maximiser.  This is the
justification of the stopping rule in the main text.  Each additional
sample reweights the calibration curve towards its falling side.

For finiteness, strict decrease of $Q$ on $(u_1,1)$ forces
$u_\star\le u_1$, because $Q$ cannot be both nondecreasing and strictly
decreasing on $(u_1,u_\star)$.  Pick $u_1<v<v'<1$.  Then
$A_N\le\mathrm{TV}(Q)$, while
$B_N\ge(v/u_\star)^N(1-v')\,|\mathrm{d}Q|((v,v'))\to\infty$ because
$|\mathrm{d}Q|((v,v'))=Q(v^+)-Q(v'^-)>0$ and $v>u_\star$.  So
$\Delta V_N<0$ for all large $N$ and $N_\star<\infty$.

\paragraph{On heterogeneity.}
Averaging single-peaked per-prompt curves whose peaks lie at different
quantiles can produce an averaged $\bar Q$ with several turning points,
for instance when the prompts split into an easy group peaking early and a
hard group peaking late.  Theorem~\ref{thm:shape} then applies to $\bar Q$,
which is the curve that the aggregate BoN value reads
(Appendix~\ref{app:proof-quantile}).  Per-prompt or per-group rank
profiles are the right tool when the averaged profile is not single-peaked.
\hfill$\square$

\subsection{Proof of Corollary~\ref{prop:return}, and the speed of return}
\label{app:proof-return}

\begin{corollary}[Return-to-prior certificate]
\label{prop:return}
If $Q$ is bounded, $\lim_{u\uparrow1}Q(u)=\mu$, and $V_{N_0}>\mu$ for some
$N_0$, then $\{V_N:N\in\mathbb N\}$ has a maximizer at finite $N$.
\end{corollary}

Once the top of $Q$ returns to the prompt mean, more samples are no longer
a safe default (Appendix~\ref{app:proof-return}).

Since $\Pr(U_{(N)}\le1-\eta)=(1-\eta)^N\to0$ for every $\eta>0$ and
$U_{(N)}$ is nondecreasing in $N$, $U_{(N)}\uparrow1$ almost surely.  As
$Q(u)\to\mu$ for $u\uparrow1$, $Q(U_{(N)})\to\mu$ almost surely, and bounded
convergence gives $V_N=\E[Q(U_{(N)})]\to\mu$.  Choose $N_1$ with
$V_N<(\mu+V_{N_0})/2$ for all $N>N_1$.  Then
$\sup_NV_N=\max_{N\le\max(N_0,N_1)}V_N$, a maximum over a finite set, which
is attained.  \hfill$\square$

\paragraph{How fast the return happens.}
Writing $W=1-U_{(N)}\sim\mathrm{Beta}(1,N)$,
\begin{equation}
\label{eq:tail-exponential-limit}
 V_N-\mu=\E\big[E(1-W)\big],\qquad NW\ \xrightarrow{d}\ \mathrm{Exp}(1).
\end{equation}
So $V_N-\mu$ is the excess of the rank profile at quantile
$1-\Theta(1/N)$, averaged with nearly exponential weights.  If
$w\mapsto E(1-w)$ is regularly varying at $0$ with index $\alpha\ge0$, the
Abelian argument of Appendix~\ref{app:proof-rates}(i) gives
\begin{equation}
\label{eq:tail-abelian-rate}
 V_N-\mu\sim\Gamma(1+\alpha)\,E(1-1/N).
\end{equation}
The speed of the return is therefore read directly off the top of $E(u)$.
By~\eqref{eq:ibp} with $Q(1^-)=\mu$, we also have
$V_N-\mu=\int u^N\,\mathrm{d}(-Q)(u)$: at large $N$, BoN beats random
selection exactly when the top of $Q$ approaches $\mu$ from above.

\subsection{Additive noise: the tilt representation and Proposition~\ref{prop:cov}}
\label{app:proof-cov}

Throughout Appendices~\ref{app:proof-cov}--\ref{app:proof-rates},
$S=R+\eps$ with $R\perp\eps$, $P_R$ is the law of $R$ on
$[r_{\min},r_{\max}]$, $\bar r=\max(|r_{\min}|,|r_{\max}|)$, and $p_\eps>0$
has a locally absolutely continuous logarithm.  Then
$p_S(s)=\int p_\eps(s-r)\,P_R(\mathrm{d}r)>0$ for every $s$,
$P(R\in\mathrm{d}r\mid S=s)=p_\eps(s-r)P_R(\mathrm{d}r)/p_S(s)$ is a
regular posterior, $G$ is continuous, and $Q=G\circ F_S^{-1}$ is a genuine
continuous version.

Then
\begin{equation}
\label{eq:posterior-mean}
 G(s)=\frac{\int r\,p_R(r)\,p_\eps(s-r)\,\mathrm{d}r}
           {\int p_R(r)\,p_\eps(s-r)\,\mathrm{d}r}.
\end{equation}

Since
$p_\eps(s-r)/p_\eps(s)=\exp\big(\int_{s-r}^{s}\psi_\eps\big)$, the posterior
at $s$ is the prior tilted by the average hazard slope over $[s-r,s]$
(Lemma~\ref{lem:tilt}); for binary reward,
exactly
\begin{equation}
\label{eq:binary-tilt}
 \operatorname{logit}G(s)=\operatorname{logit}\mu+\int_{s-1}^{s}\psi_\eps(t)\,\mathrm{d}t,
 \qquad R\in\{0,1\}.
\end{equation}

\begin{lemma}[Tilt representation]
\label{lem:tilt}
For all $s,r$, $p_\eps(s-r)/p_\eps(s)=\exp\big(\int_{s-r}^{s}\psi_\eps(t)\,\mathrm{d}t\big)$,
with the convention $\int_a^b=-\int_b^a$.  Consequently
\begin{equation}
\label{eq:posterior-tilt}
 G(s)=\frac{\E[R\,w_s(R)]}{\E[w_s(R)]},\qquad
 w_s(r)=\exp\Big(\int_{s-r}^{s}\psi_\eps(t)\,\mathrm{d}t\Big).
\end{equation}
If $R\in\{a,b\}$ with $a<b$ and $\Pr(R=b)=\mu_b\in(0,1)$, then
\begin{equation}
\label{eq:two-point}
 \operatorname{logit}\Pr(R=b\mid S=s)=\operatorname{logit}\mu_b
 +\int_{s-b}^{s-a}\psi_\eps(t)\,\mathrm{d}t ,
\end{equation}
and~\eqref{eq:binary-tilt} is the case $a=0$, $b=1$.
\end{lemma}

\begin{proof}
Since $\log p_\eps$ is locally absolutely continuous with derivative
$-\psi_\eps$,
$\log p_\eps(s)-\log p_\eps(s-r)=-\int_{s-r}^{s}\psi_\eps$.  Dividing the
numerator and denominator of $G(s)$ by $p_\eps(s)$ gives the second claim.
For~\eqref{eq:two-point}, the posterior odds are the prior odds times
$p_\eps(s-b)/p_\eps(s-a)$, and
$\log p_\eps(s-b)-\log p_\eps(s-a)=\int_{s-b}^{s-a}\psi_\eps$.
\end{proof}

The weight $w_s$ is the exponential tilt $e^{\lambda r}$ with the constant
rate $\lambda$ replaced by the average hazard slope over the window between
$s-r$ and $s$.  When $\psi_\eps$ is nearly constant on windows of length
$\bar r$, which is the situation at large $|s|$ under a convergent hazard,
$G(s)\approx m(\psi_\eps(s))$.  Every result in
Sections~\ref{sec:additive} and~\ref{sec:where} is a statement about how
this effective rate behaves.

\paragraph{Properties of $m$.}
With $M(\lambda)=\E e^{\lambda R}$ and the tilted law
$P_\lambda(\mathrm{d}r)\propto e^{\lambda r}P_R(\mathrm{d}r)$, we have
$m=(\log M)'$ and $m'(\lambda)=\operatorname{Var}_{P_\lambda}(R)\ge0$, with
strict inequality unless $R$ is degenerate; also $m(0)=\mu$.  If $r_{\max}$
is in the support of $R$, then for every $\delta>0$
\begin{equation}
\label{eq:tilt-concentration}
 \frac{P_\lambda(R\le r_{\max}-\delta)}{P_\lambda(R>r_{\max}-\delta/2)}
 \le\frac{e^{-\lambda\delta/2}}{\Pr(R>r_{\max}-\delta/2)}\to0,
\end{equation}
so $m(\lambda)\to r_{\max}$, and $m(\lambda)<r_{\max}$ for finite $\lambda$
unless $R\equiv r_{\max}$.  For binary $R$, \eqref{eq:two-point} with
$\psi_\eps\equiv\lambda$ gives
$\operatorname{logit}m(\lambda)=\operatorname{logit}\mu+\lambda$.

\begin{proposition}[Covariance identity]
\label{prop:cov}
Whenever differentiation under the integral is permitted (e.g.\
$\psi_\eps$ continuous),
\begin{equation}
\label{eq:covariance-identity}
 G'(s)=-\operatorname{Cov}\big(R,\;\psi_\eps(s-R)\;\big|\;S=s\big).
\end{equation}
Hence if $p_\eps$ is log-concave, $G$ and $Q$ are nondecreasing and, by
Theorem~\ref{thm:shape}(a), so is $V_N$ for \emph{every} prior $p_R$.  For
Gaussian error, $G'(s)=\operatorname{Var}(R\mid S=s)/\sigma^2$.
\end{proposition}

\paragraph{Proof of the covariance identity~\eqref{eq:covariance-identity}.}
Assume $\psi_\eps$ is continuous.  For $s$ in a compact interval $J$ and
$r\in[r_{\min},r_{\max}]$, the derivative
$\partial_sp_\eps(s-r)=-\psi_\eps(s-r)p_\eps(s-r)$ is bounded, so
differentiation under the integral is permitted in both the numerator and
the denominator of $G$.  First,
$p_S'(s)=-\int\psi_\eps(s-r)p_\eps(s-r)\,P_R(\mathrm{d}r)$, so the hazard of
the score is the posterior mean of the hazard of the error:
$\psi_S(s):=-(\log p_S)'(s)=\E[\psi_\eps(s-R)\mid S=s]$.  Then
\begin{equation}
\label{eq:covariance-proof}
\begin{aligned}
 G'(s)&=-\E\big[R\,\psi_\eps(s-R)\mid S=s\big]
       +\E[R\mid S=s]\,\E\big[\psi_\eps(s-R)\mid S=s\big] \\
      &=-\operatorname{Cov}\big(R,\psi_\eps(s-R)\mid S=s\big).
\end{aligned}
\end{equation}
If $\psi_\eps$ is nondecreasing, $r\mapsto\psi_\eps(s-r)$ is nonincreasing,
and Chebyshev's association inequality gives a nonpositive covariance: for
$X'$ an independent copy of $X$, $f$ nondecreasing and $g$ nonincreasing,
$\operatorname{Cov}(f(X),g(X))=\tfrac12\E\big[(f(X)-f(X'))(g(X)-g(X'))\big]\le0$
because the integrand is pointwise nonpositive.  For Gaussian error,
$\psi_\eps(s-R)=(s-R)/\sigma^2$ and the covariance is
$-\operatorname{Var}(R\mid S=s)/\sigma^2$.  Finally $F_S^{-1}$ is
increasing, so $Q$ inherits the monotonicity of $G$.

The same computation localises the failure.  $G'(s)<0$ requires
$r\mapsto\psi_\eps(s-r)$ to fail to be nonincreasing on
$[r_{\min},r_{\max}]$, that is, $\psi_\eps$ must decrease somewhere on
$[s-r_{\max},s-r_{\min}]$.  For a two-point prior, \eqref{eq:two-point}
makes this exact:
\begin{equation}
\label{eq:two-point-derivative}
 G'(s)=(b-a)\,\pi_s(1-\pi_s)\,\big[\psi_\eps(s-a)-\psi_\eps(s-b)\big],
 \qquad \pi_s=\Pr(R=b\mid S=s).
\end{equation}
The calibration curve falls exactly where the larger error $s-a$, which
explains $s$ by the low reward, is cheaper per unit than the smaller error
$s-b$, which explains it by the high reward.

\paragraph{Sharpness, without differentiability.}

\begin{proposition}[Log-concavity is exactly the monotonicity condition]
\label{prop:lc-sharp}
For $p_\eps$ as above the following are equivalent:
\begin{enumerate}
\item[(a)] $p_\eps$ is log-concave;
\item[(b)] for every law of $R$ on a bounded interval, the posterior
$P(R\in\cdot\mid S=s)$ is nondecreasing in $s$ in the likelihood-ratio
order; in particular $\E[\varphi(R)\mid S=s]$ is nondecreasing in $s$ for
every bounded nondecreasing $\varphi$, and $G$ is nondecreasing;
\item[(c)] $G$ is nondecreasing for every two-point law of $R$.
\end{enumerate}
\end{proposition}

\begin{proof}
(a)$\Rightarrow$(b).  Let $s<s'$ and $f=\log p_\eps$.  The posterior at
$s'$ has density proportional to
$\rho(r)=p_\eps(s'-r)/p_\eps(s-r)$ with respect to the posterior at $s$,
and $\log\rho(r)=f(x+d)-f(x)$ with $x=s-r$ and $d=s'-s>0$.  For concave
$f$ the increment $f(x+d)-f(x)$ is nonincreasing in $x$; as $r$ increases,
$x$ decreases, so $\rho$ is nondecreasing in $r$.  This is the
likelihood-ratio order.  With $\tilde\rho=\rho/\E_s[\rho]$ (mean one under
the posterior at $s$), $\E_{s'}[\varphi(R)]-\E_s[\varphi(R)]
=\operatorname{Cov}_s(\varphi(R),\tilde\rho(R))\ge0$ by Chebyshev's
inequality with two nondecreasing functions.
(b)$\Rightarrow$(c) is immediate.
(c)$\Rightarrow$(a).  Take $R\in\{a,b\}$ with $a<b$.
By~\eqref{eq:two-point}, $G$ is nondecreasing if and only if
$s\mapsto f(s-b)-f(s-a)$ is nondecreasing.  Writing $x=s-b$ and $d=b-a$,
this says that $x\mapsto f(x+d)-f(x)$ is nonincreasing, for every $d>0$
since $a<b$ are arbitrary.  Comparing $x$ with $x+d$ gives
$f(x+d)-f(x)\ge f(x+2d)-f(x+d)$, i.e.\ $f(x+d)\ge\tfrac12\big(f(x)+f(x+2d)\big)$.
Thus $f$ is midpoint concave, and being continuous it is concave.
\end{proof}

The equivalence has two readings.  First, the failure mechanism identified
in Section~\ref{sec:additive} is the exact boundary and not merely a
convenient sufficient condition: for every non-log-concave error there is a
binary reward whose calibration curve decreases somewhere.  Second,
log-concavity of $p_\eps$ is equivalent to total positivity of order two of
the location kernel $(s,r)\mapsto p_\eps(s-r)$ \citep{Karlin1968}.  The
total positivity that drives Theorem~\ref{thm:shape} on the side of $N$ is
therefore the same property that drives monotonicity of $G$ on the side of
the score.  Under (b) the conclusion holds for $\E[\varphi(R)\mid S=s]$ with
any nondecreasing $\varphi$, so it does not depend on the scale on which
utility is measured.  \hfill$\square$

\subsection{Proof of Theorem~\ref{thm:hazard}}
\label{app:proof-hazard}

\paragraph{Step 0: $\lambda_\star\ge0$, and $\lambda_\star$ is a
log-survival slope.}
If $\lambda_\star<0$, then $\psi_\eps\le\lambda_\star/2$ on some
$[M,\infty)$, so $\log p_\eps(t)\ge\log p_\eps(M)+|\lambda_\star|(t-M)/2$,
which contradicts integrability.  Integrability together with the
existence of $\lambda_\star$ also forces $p_\eps(t)\to0$.  For
$\lambda_\star>0$ this is clear.  For $\lambda_\star=0$, note that
$p_\eps(t+h)/p_\eps(t)=\exp(-\int_t^{t+h}\psi_\eps)\to1$ for every fixed
$h$; if $p_\eps(t_n)\ge c>0$ along $t_n\to\infty$, then $p_\eps\ge c/2$ on
infinitely many disjoint intervals of fixed length, which is impossible.
L'H\^opital's rule applied to $p_\eps/\bar F_\eps$ then gives
\begin{equation}
\label{eq:survival-slope}
 -\frac{\mathrm{d}}{\mathrm{d}t}\log\bar F_\eps(t)=\frac{p_\eps(t)}{\bar F_\eps(t)}
 \;\longrightarrow\;\lim_{t\to\infty}\frac{-p_\eps'(t)}{p_\eps(t)}=\lambda_\star .
\end{equation}
This is why $\lambda_\star$ can be estimated from a log-survival plot, as
suggested in the main text.

\paragraph{Step 1: $\lambda_\star<\infty$.}
Let $\eta_M=\operatorname*{ess\,sup}_{t\ge M}|\psi_\eps(t)-\lambda_\star|$,
so $\eta_M\to0$.  For $|r|\le\bar r$ the interval between $s-r$ and $s$ lies
in $[s-\bar r,s+\bar r]$, so for $s\ge M+\bar r$
\begin{equation}
\label{eq:finite-hazard-bound}
 \Big|\int_{s-r}^{s}\psi_\eps-\lambda_\star r\Big|\le\bar r\,\eta_M .
\end{equation}
Hence $w_s(r)\to e^{\lambda_\star r}$ uniformly in $r\in[r_{\min},r_{\max}]$,
and $w_s$ is bounded above and away from zero uniformly for large $s$.
Lemma~\ref{lem:tilt} and bounded convergence give
$G(s)\to\E[Re^{\lambda_\star R}]/\E[e^{\lambda_\star R}]=m(\lambda_\star)$.
If $R$ is nondegenerate, the properties of $m$ give
$\mu<m(\lambda_\star)<r_{\max}$ for $\lambda_\star\in(0,\infty)$.

\paragraph{Step 2: $\lambda_\star=\infty$.}
Fix $\delta>0$ and let $A=[r_{\min},r_{\max}-\delta]$ and
$B=(r_{\max}-\delta/2,r_{\max}]$.  For $r\in A$ and $r'\in B$ we have
$r'-r\ge\delta/2$ and $s-r'\ge s-r_{\max}$, so
\begin{equation}
\label{eq:divergent-hazard-ratio}
 \frac{p_\eps(s-r)}{p_\eps(s-r')}
 =\exp\Big(-\int_{s-r'}^{s-r}\psi_\eps\Big)
 \le\exp\Big(-\tfrac{\delta}{2}\,\operatorname*{ess\,inf}_{t\ge s-r_{\max}}\psi_\eps(t)\Big)
 \;\longrightarrow\;0 .
\end{equation}
Integrating over $A$ and $B$,
$\Pr(R\in A\mid S=s)/\Pr(R\in B\mid S=s)\le
\big[\Pr(R\in A)/\Pr(R\in B)\big]\cdot o(1)\to0$, where $\Pr(R\in B)>0$ by
the support assumption.  Hence
$G(s)\ge(r_{\max}-\delta)-(r_{\max}-r_{\min})\Pr(R\in A\mid S=s)\to
r_{\max}-\delta$; since $\delta$ is arbitrary and $G\le r_{\max}$,
$G(s)\to r_{\max}$.

\paragraph{Step 3: transfer to $Q$.}
$p_S>0$ on $\mathbb R$, so $F_S$ is a continuous increasing bijection onto
$(0,1)$ and $F_S^{-1}(u)\to\infty$ as $u\uparrow1$.

\paragraph{Step 4: the examples.}
Gaussian: $\psi_\eps(t)=t/\sigma^2\to\infty$.  Generalised Gaussian
$p_\eps\propto e^{-|t|^\beta}$ (log-density locally absolutely continuous
for every $\beta>0$): $\psi_\eps(t)=\beta t^{\beta-1}$ for $t>0$, which
tends to $\infty$, $1$ or $0$ as $\beta>1$, $=1$ or $<1$.  Laplace with
scale $b$: $\psi_\eps=\operatorname{sign}(t)/b$.  Logistic with scale $b$:
$\psi_\eps(t)=\tanh(t/(2b))/b\to1/b$.  Tails $\propto t^ke^{-\lambda t}$:
$\psi_\eps(t)=\lambda-k/t\to\lambda$.  Student-$t_\nu$:
$\psi_\eps(t)=(\nu+1)t/(\nu+t^2)\sim(\nu+1)/t$.  Pareto-type tails
$\propto t^{-(\alpha+1)}$: $\psi_\eps(t)=(\alpha+1)/t$.  Log-normal-type
tails: $\psi_\eps(t)=\big(1+(\log t-m)/s^2\big)/t\to0$.

\paragraph{The mechanism: explaining away.}
Steps~1 and~2 are a likelihood-ratio contest between explanations of the
same extreme score.  Crediting one more unit of reward to the response
saves one unit of error, which is worth about $\psi_\eps(s)$ nats of
likelihood.  If this saving grows without bound, only the highest reward is
a credible explanation; if it converges to $\lambda_\star$, every reward
level keeps a posterior weight proportional to $e^{\lambda_\star r}$; if it
vanishes, the score carries no information about which reward produced it
and the posterior falls back to the prior.  By~\eqref{eq:binary-tilt}, in
the binary case the log-odds lift at the top of the ranking is exactly
$\int_{s-1}^{s}\psi_\eps\to\lambda_\star$.  \hfill$\square$

\subsection{Rates of approach: Proposition~\ref{prop:rates}}
\label{app:proof-rates}

\begin{proposition}[Rates of approach]
\label{prop:rates}
\begin{enumerate}
\item[(i)] \textup{(Heavy tails)} If $p_\eps$ is regularly varying at
$+\infty$ with index $-(\nu+1)$, $\nu>0$, and $\psi_\eps(t)\sim(\nu+1)/t$,
then $G(s)-\mu=(\nu+1)\operatorname{Var}(R)\,s^{-1}(1+o(1))$ and
\begin{equation}
\label{eq:heavy-tail-rate}
 V_N-\mu\;\sim\;\frac{(\nu+1)\,\Gamma(1+1/\nu)\,\operatorname{Var}(R)}{s_N},
 \qquad s_N:=\bar F_\eps^{-1}(1/N)\asymp N^{1/\nu}\ell(N),
\end{equation}
with $\ell$ slowly varying: heavier tails return to the prior faster.
\item[(ii)] \textup{(Exponential tails)} If $p_\eps(t)\propto e^{-\lambda_\star t}$
for $t>t_0$, then $G(s)=m(\lambda_\star)$ for $s>t_0+r_{\max}$ and
$|V_N-m(\lambda_\star)|\le(r_{\max}-r_{\min})\,F_S(t_0+r_{\max})^{N}$.
\item[(iii)] \textup{(Gaussian tails, binary reward)} If
$\eps\sim\mathcal N(0,\sigma^2)$ and $R\in\{0,1\}$, $\mu\in(0,1)$, then
\begin{equation}
\label{eq:gaussian-binary-rate}
 1-V_N=\exp\!\Big(-(1+o(1))\,\sqrt{2\log N}\,\big/\,\sigma\Big),
\end{equation}
slower than any power of $N$.
\item[(iv)] \textup{(Gaussian tails, continuous reward)} If
$\eps\sim\mathcal N(0,\sigma^2)$ and $R$ has density
$\asymp(r_{\max}-r)^{\beta}$ near $r_{\max}$, then
$r_{\max}-V_N\sim(\beta+1)\,\sigma/\sqrt{2\log N}$.
\end{enumerate}
\end{proposition}

All four follow from $G(s_N)\approx m\big(\psi_\eps(s_N)\big)$
(Appendix~\ref{app:proof-rates}): extra samples are wasted beyond a
geometric horizon at a plateau, help at super-polynomial cost under
Gaussian tails, and hurt beyond the peak under heavy tails.

\paragraph{(i) Heavy tails.}
Assume $\operatorname{Var}(R)>0$; otherwise both sides vanish.

\emph{Step 1: calibration.}  Since $t\,\psi_\eps(t)\to\nu+1$,
$\psi_\eps(t)=(\nu+1)/s+o(1/s)$ uniformly for $t\in[s-\bar r,s+\bar r]$, so
$\int_{s-r}^{s}\psi_\eps=(\nu+1)r/s+o(1/s)$ uniformly in $|r|\le\bar r$ and
$w_s(r)=1+(\nu+1)r/s+o(1/s)$.  By Lemma~\ref{lem:tilt},
\begin{equation}
\label{eq:heavy-tail-calibration}
 G(s)=\frac{\mu+(\nu+1)\E[R^2]/s+o(1/s)}{1+(\nu+1)\mu/s+o(1/s)}
     =\mu+\frac{(\nu+1)\operatorname{Var}(R)}{s}+o(1/s).
\end{equation}
Equivalently $G(s)=m(\psi_\eps(s))+o(1/s)$, because
$m(\lambda)=\mu+\operatorname{Var}(R)\lambda+O(\lambda^2)$: at extreme scores
the posterior is the prior tilted by the local hazard slope.

\emph{Step 2: the tail of $S$.}  By Karamata's theorem
\citep[Thm.~1.5.11]{Bingham1987}, $\bar F_\eps$ is regularly varying with
index $-\nu$.  By the uniform convergence theorem for regularly varying
functions, $\bar F_\eps(s-r)/\bar F_\eps(s)\to1$ uniformly in $|r|\le\bar r$,
so $\bar F_S(s)=\E[\bar F_\eps(s-R)]\sim\bar F_\eps(s)$.  Asymptotically
equivalent regularly varying functions of negative index have
asymptotically equivalent inverses \citep[Thm.~1.5.12]{Bingham1987}, so
$\bar F_S^{-1}(w)\sim\bar F_\eps^{-1}(w)=w^{-1/\nu}/L(w)$ as $w\downarrow0$,
with $L$ slowly varying at $0$.

\emph{Step 3: the rank profile.}  Combining Steps~1 and~2,
\begin{equation}
\label{eq:heavy-tail-rank-profile}
 E(1-w)=G\big(\bar F_S^{-1}(w)\big)-\mu
 =(\nu+1)\operatorname{Var}(R)\,w^{1/\nu}L(w)\,(1+o(1)),\qquad w\downarrow0 .
\end{equation}

\emph{Step 4: the Abelian step.}  Let $W=1-U_{(N)}$, with density
$N(1-w)^{N-1}$.  Fix a small $\eta>0$.  The event $\{W>\eta\}$ contributes
at most $\|E\|_\infty(1-\eta)^N$, which is exponentially small.  On
$\{W\le\eta\}$ substitute $w=x/N$:
\begin{equation}
\label{eq:abelian-integral}
 \E\big[W^{1/\nu}L(W);\,W\le\eta\big]
 =N^{-1/\nu}L(1/N)\int_0^{\eta N}x^{1/\nu}\,\frac{L(x/N)}{L(1/N)}\,
  \Big(1-\frac{x}{N}\Big)^{N-1}\mathrm{d}x .
\end{equation}
Pointwise, $L(x/N)/L(1/N)\to1$ and $(1-x/N)^{N-1}\to e^{-x}$.  Potter's
bounds \citep[Thm.~1.5.6]{Bingham1987} give, for any $\gamma>0$ and $\eta$
small enough, $L(x/N)/L(1/N)\le2\max(x^\gamma,x^{-\gamma})$, and
$(1-x/N)^{N-1}\le e^{-x/2}$ for $N\ge2$.  Taking $\gamma<1/\nu$, dominated
convergence yields $\int_0^\infty x^{1/\nu}e^{-x}\,\mathrm{d}x=\Gamma(1+1/\nu)$.
Because the factor $1+o(1)$ in Step~3 is uniform on $w\le\eta$, we obtain
\begin{equation}
\label{eq:abelian-limit}
\begin{aligned}
 V_N-\mu=\E[E(1-W)]
 &\sim(\nu+1)\Gamma(1+1/\nu)\operatorname{Var}(R)\,N^{-1/\nu}L(1/N) \\
 &=\frac{(\nu+1)\Gamma(1+1/\nu)\operatorname{Var}(R)}{s_N},
\end{aligned}
\end{equation}
since $s_N=\bar F_\eps^{-1}(1/N)=N^{1/\nu}/L(1/N)$.  The constant is
accurate at moderate budgets.  For Student-$t_3$ noise and $R\in\{0,1\}$
with $\mu=1/2$, direct evaluation of Equation~\eqref{eq:bon-quantile}
gives ratios of $V_N-\mu$ to the asymptotic expression of $0.81$, $0.95$,
$0.99$ and $0.998$ at $N=10^2,10^3,10^4,10^5$.

The exponent $1/\nu$ appears for a transparent reason: the budget-$N$
winner sits at score $s_N\asymp N^{1/\nu}$, and there the local tilt is
$(\nu+1)/s_N$.  Heavier tails move the winner out faster and so make the
tilt vanish sooner.

\paragraph{(ii) Exponential tails.}
For $s>t_0+r_{\max}$ we have $s-r>t_0$ for every $r$ in the support, so
$p_\eps(s-r)=c\,e^{-\lambda_\star(s-r)}$ and the posterior is exactly
$\propto e^{\lambda_\star r}P_R(\mathrm{d}r)$, i.e.\
$G(s)=m(\lambda_\star)$.  Since $V_N=\E[G(S_{(N)})]$ and both $G$ and $m$
take values in $[r_{\min},r_{\max}]$,
\begin{equation}
\label{eq:exponential-rate}
 |V_N-m(\lambda_\star)|\le(r_{\max}-r_{\min})\Pr\big(S_{(N)}\le t_0+r_{\max}\big)
 =(r_{\max}-r_{\min})\,F_S(t_0+r_{\max})^N ,
\end{equation}
with $F_S(t_0+r_{\max})<1$ because $p_S>0$.  Solving
$(r_{\max}-r_{\min})F_S(t_0+r_{\max})^N\le\delta$ for $N$ gives the
horizon stated in the main text.

\paragraph{(iii) Gaussian tails, binary reward.}
By~\eqref{eq:binary-tilt} with $\psi_\eps(t)=t/\sigma^2$,
$\int_{s-1}^{s}t\,\mathrm{d}t/\sigma^2=(2s-1)/(2\sigma^2)$, so exactly
\begin{equation}
\label{eq:gaussian-binary-calibration}
 1-G(s)=\Big(1+\tfrac{\mu}{1-\mu}\,e^{(2s-1)/(2\sigma^2)}\Big)^{-1},
 \qquad -\log\big(1-G(s)\big)=\frac{s}{\sigma^2}\,(1+o(1)).
\end{equation}
Let $a_N=\bar F_S^{-1}(1/N)$.  Since
$\bar F_S(s)=(1-\mu)\bar\Phi(s/\sigma)+\mu\bar\Phi((s-1)/\sigma)$ and
$-\log\bar\Phi(z)=\tfrac{z^2}{2}(1+o(1))$, we have
$-\log\bar F_S(s)=\tfrac{s^2}{2\sigma^2}(1+o(1))$ and
$a_N=\sigma\sqrt{2\log N}\,(1+o(1))$.  $G$ is nondecreasing
(Proposition~\ref{prop:cov}) and $0\le1-G\le1$.  Fix $\delta\in(0,1)$.

\emph{Upper bound.}  Splitting on $\{S_{(N)}\le(1-\delta)a_N\}$,
\begin{equation}
\label{eq:gaussian-upper-bound}
 1-V_N\le1-G\big((1-\delta)a_N\big)+\Pr\big(S_{(N)}\le(1-\delta)a_N\big).
\end{equation}
The first term is $\exp\big(-(1-\delta)(1+o(1))\sqrt{2\log N}/\sigma\big)$.
For the second, $\Pr(S_{(N)}\le x)=F_S(x)^N\le\exp(-N\bar F_S(x))$ and
$N\bar F_S((1-\delta)a_N)=N^{1-(1-\delta)^2+o(1)}$, which grows polynomially.
The second term is therefore $\exp(-N^{c})$ for some $c>0$, which is
negligible.

\emph{Lower bound.}  Restricting to $\{S_{(N)}\le(1+\delta)a_N\}$,
\begin{equation}
\label{eq:gaussian-lower-bound}
 1-V_N\ge\big[1-G\big((1+\delta)a_N\big)\big]\,\Pr\big(S_{(N)}\le(1+\delta)a_N\big),
\end{equation}
and $\Pr(S_{(N)}\le(1+\delta)a_N)\ge1-N\bar F_S((1+\delta)a_N)
=1-N^{1-(1+\delta)^2+o(1)}\to1$.

Letting $\delta\downarrow0$ gives
$-\log(1-V_N)=(1+o(1))\sqrt{2\log N}/\sigma$.  The rate is slow because
the tilt at the winning score is $a_N/\sigma^2$, and light tails make
$a_N$ grow only like $\sqrt{\log N}$: raising the top score by one unit
requires multiplying $N$ by roughly $e^{a_N/\sigma^2}$.

\paragraph{(iv) Gaussian tails, continuous reward.}
Assume $P_R(\mathrm{d}r)=p_R(r)\,\mathrm{d}r$ with
$p_R(r)=c\,(r_{\max}-r)^\beta(1+o(1))$ as $r\uparrow r_{\max}$, $\beta>-1$.
For Gaussian error, $p_\eps(s-r)\propto e^{-s^2/(2\sigma^2)}\,
e^{-r^2/(2\sigma^2)}\,e^{(s/\sigma^2)r}$, so $G(s)=\tilde m(s/\sigma^2)$
\emph{exactly}, where $\tilde m$ is the tilt function of the modified prior
$\tilde p_R\propto p_R\,e^{-r^2/(2\sigma^2)}$.  This prior has the same
$(r_{\max}-r)^\beta$ behaviour at the top.  Under the tilted law the gap
$y=r_{\max}-R$ has density $\propto y^\beta(1+o(1))e^{-\lambda y}$ near
$y=0$, so by Laplace's method it is asymptotically
$\mathrm{Gamma}(\beta+1,\lambda)$ and $r_{\max}-\tilde m(\lambda)\sim(\beta+1)/\lambda$.
Hence $r_{\max}-G(s)\sim(\beta+1)\sigma^2/s$.  The two-sided argument
of~(iii), with $a_N=\sigma\sqrt{2\log N}(1+o(1))$ unchanged because $R$ is
bounded, gives
$r_{\max}-V_N\sim(\beta+1)\sigma^2/a_N=(\beta+1)\sigma/\sqrt{2\log N}$.
\hfill$\square$

\subsection{Proof of Proposition~\ref{prop:scale} and of Proposition~\ref{prop:binaryLR}}
\label{app:proof-scale}

\begin{proposition}[Binary reward: the tail likelihood ratio decides]
\label{prop:binaryLR}
Let $R=C\in\{0,1\}$ and $p_1,p_0$ the densities of $S$ given $C=1,0$, with
no further assumption.  Then
\begin{equation}
\label{eq:binary-likelihood-ratio}
 \operatorname{logit}Q(u)=\operatorname{logit}\mu+\log\Lambda\big(F_S^{-1}(u)\big),
 \qquad \Lambda=p_1/p_0 ,
\end{equation}
so $\lim_{u\uparrow1}Q(u)$ is $1$, $\mu$, or $0$ as $\Lambda(s)\to\infty$,
$1$, or $0$, and a plateau in $(0,1)$ for any other finite positive limit.
\end{proposition}

Theorem~\ref{thm:hazard} is the case
$\log\Lambda(s)=\int_{s-1}^{s}\psi_\eps$ of~\eqref{eq:binary-tilt},
Proposition~\ref{prop:scale} that of Gaussian classes with unequal
variances (Appendix~\ref{app:proof-scale}); in verifiable tasks $\Lambda$
at the top of the score range can be estimated directly.

Write $\varphi_\sigma(x)=(2\pi\sigma^2)^{-1/2}e^{-x^2/(2\sigma^2)}$ (and
$\varphi_0\equiv0$ off the origin), $\sigma_{\max}=\sigma(r_\sigma)$, and
$P(R\in\mathrm{d}r\mid S=s)\propto\varphi_{\sigma(r)}(s-r)\,P_R(\mathrm{d}r)$.

\paragraph{A monotonicity fact.}
For fixed $x\ne0$,
$\partial_\sigma\log\varphi_\sigma(x)=-1/\sigma+x^2/\sigma^3$, which is
positive for $\sigma<|x|$.  So when the discrepancy $|x|$ exceeds every
available scale, a wider conditional law always assigns it more density.
We take $s\ge2\bar r+2\sigma_{\max}$, so that
$|s-r|\ge s-\bar r>\sigma_{\max}$ for all $r$.

\paragraph{Concentration at the noisiest level.}
Fix $\eta\in(0,\sigma_{\max})$ and put
$A_\eta=\{r:\sigma(r)\le\sigma_{\max}-\eta\}$ and
$D_\eta=\{r:\sigma(r)\ge\sigma_{\max}-\eta/2\}$.  By continuity $D_\eta$
contains a neighbourhood of $r_\sigma$ in $[r_{\min},r_{\max}]$, so
$\Pr(R\in D_\eta)>0$ because $r_\sigma$ is in the support.  For
$r\in A_\eta$, monotonicity in $\sigma$ and then in $|x|$ gives
\begin{equation}
\label{eq:scale-upper-density}
 \varphi_{\sigma(r)}(s-r)\le\varphi_{\sigma_{\max}-\eta}(s-r)
 \le\frac{1}{\sqrt{2\pi}(\sigma_{\max}-\eta)}
 \exp\Big(-\frac{(s-\bar r)^2}{2(\sigma_{\max}-\eta)^2}\Big),
\end{equation}
and this bound needs no lower bound on $\sigma(r)$.  For $r\in D_\eta$,
likewise,
\begin{equation}
\label{eq:scale-lower-density}
 \varphi_{\sigma(r)}(s-r)\ge\varphi_{\sigma_{\max}-\eta/2}(s+\bar r)
 =\frac{1}{\sqrt{2\pi}(\sigma_{\max}-\eta/2)}
 \exp\Big(-\frac{(s+\bar r)^2}{2(\sigma_{\max}-\eta/2)^2}\Big).
\end{equation}
The ratio of the two bounds is $\exp(-c_\eta s^2+O(s))$ with
$c_\eta=\tfrac12\big[(\sigma_{\max}-\eta)^{-2}-(\sigma_{\max}-\eta/2)^{-2}\big]>0$,
so $\Pr(R\in A_\eta\mid S=s)/\Pr(R\in D_\eta\mid S=s)\to0$.  Hence
$\Pr(\sigma(R)>\sigma_{\max}-\eta\mid S=s)\to1$ for every $\eta>0$.

\paragraph{From scale to location.}
Because $\sigma$ is continuous on the compact interval $[r_{\min},r_{\max}]$
with a unique maximiser, for every $\delta>0$ there is $\eta>0$ with
$\{\sigma>\sigma_{\max}-\eta\}\subset(r_\sigma-\delta,r_\sigma+\delta)$.
Otherwise some sequence $r_n$ with $\sigma(r_n)\to\sigma_{\max}$ and
$|r_n-r_\sigma|\ge\delta$ would have a limit point that is a second
maximiser.  So the posterior concentrates at $r_\sigma$ and
$G(s)\to r_\sigma$.  Finally $V_N=\E[G(S_{(N)})]$, $S_{(N)}\to\infty$ almost
surely because $S$ has unbounded upper support (each conditional law is
Gaussian with $\sigma(R)>0$), and bounded convergence gives
$V_N\to r_\sigma$.  \hfill$\square$

\paragraph{Why scale beats shift.}
The log-likelihood is
\begin{equation}
\label{eq:gaussian-log-likelihood}
 \log\varphi_{\sigma(r)}(s-r)
 =-\frac{s^2}{2\sigma^2(r)}+\frac{s\,r}{\sigma^2(r)}-\frac{r^2}{2\sigma^2(r)}-\log\sigma(r)+\text{const}.
\end{equation}
The order-$s^2$ term depends on $r$ only through the scale, and the reward
enters first at order $s$.  With homoskedastic noise the $s^2$ term is
common to all $r$, the order-$s$ term $sr/\sigma^2$ is maximised at
$r_{\max}$, and we recover Theorem~\ref{thm:hazard} with
$\lambda_\star=\infty$.  With heteroskedastic noise the $s^2$ term decides.

\paragraph{The multiplicative model.}
If $S=R(1+\eps)$ with $\eps\sim\mathcal N(0,\sigma_0^2)$, then
$S\mid R=r\sim\mathcal N(r,\sigma_0^2r^2)$ and $\sigma(r)=\sigma_0|r|$.  If
$\Pr(R=0)=0$, $|r_{\min}|>r_{\max}$ and $r_{\min}$ is in the support, the
unique maximiser is $r_{\min}$, and the proposition gives
$V_N\to r_{\min}$: BoN converges to the worst reward level.  The vanishing
of $\sigma$ at $r=0$ is harmless because the upper bound on $A_\eta$ does
not use a lower bound on $\sigma$.

\paragraph{Proof of Proposition~\ref{prop:binaryLR}.}
Bayes' rule gives
$\Pr(C=1\mid S=s)/\Pr(C=0\mid S=s)=\mu\,p_1(s)/\big((1-\mu)p_0(s)\big)$;
take logarithms and compose with $F_S^{-1}$.  The limits follow from
continuity of the logistic function on $[-\infty,\infty]$.  In the additive
model $p_1(s)/p_0(s)=p_\eps(s-1)/p_\eps(s)=\exp(\int_{s-1}^{s}\psi_\eps)$ by
Lemma~\ref{lem:tilt}.  For Gaussian classes with scales $\sigma_1,\sigma_0$,
$\log\Lambda(s)=s^2\big(\sigma_0^{-2}-\sigma_1^{-2}\big)/2+O(s)$, which
tends to $+\infty$ or $-\infty$ according as $\sigma_1>\sigma_0$ or
$\sigma_1<\sigma_0$.

\paragraph{The lexicographic principle for binary reward.}
Suppose $-\log p_c(s)=a_cs^{\beta_c}\big(1+o(1)\big)$ as $s\to\infty$ for
$c\in\{0,1\}$.  If $\beta_0<\beta_1$, incorrect responses have the heavier
tail, $\log\Lambda\to-\infty$ and $Q(u)\to0$, so BoN converges to the
\emph{worst} outcome even if correct responses score higher in the bulk.
If $\beta_0=\beta_1$ and $a_0<a_1$, the wider incorrect class wins in the
same way.  Only when shape and scale agree do lower-order terms, among them
the location shift, decide; in the additive model they decide through
$\int_{s-1}^{s}\psi_\eps$, which is Theorem~\ref{thm:hazard}.  An extreme
score is thus explained first by the heaviest tail, then by the largest
scale, and only then by the largest reward.

\subsection{Proofs for Section~\ref{sec:where}}
\label{app:proof-where}

\begin{proposition}[Contamination]
\label{prop:contam}
Let $p_\eps=(1-\eta)\,p_0+\eta\,p_1$, $\eta\in(0,1]$, with $p_0$ having
generalised Gaussian or exponential tails ($\beta\ge1$) and $p_1$
satisfying the condition of Proposition~\ref{prop:rates}(i).  Then
$\lambda_\star=0$ for every $\eta>0$, so $\lim_{u\uparrow1}Q(u)=\mu$ and
Corollary~\ref{prop:return} applies once a finite budget beats random.
\end{proposition}

Any heavy-tailed component, however small, eventually forces a return to
the prior; by Lemma~\ref{lem:invisible}, smaller $\eta$ only delays the
turn (Appendix~\ref{app:proof-where}).

\begin{corollary}[No finite certificate of monotonicity]
\label{cor:nocert}
Let $Q$ take values in $[r_{\min},r_{\max}]$, $K\in\mathbb N$, $\gamma>0$.
Then there exist $\delta>0$ and $Q'$ with values in $[r_{\min},r_{\max}]$,
equal to $Q$ on $[0,1-\delta]$ and constant on $(1-\delta,1)$ with value
$\mu'=\int_0^1Q'$, such that $|V'_N-V_N|\le\gamma$ for all $N\le K$ and
$\lim_{N\to\infty}V'_N=\mu'$.
\end{corollary}

No BoN curve up to budget $K$ rules out a later return, so our diagnostics
target $\lambda_\star$ and $\Lambda$ directly
(Appendix~\ref{app:proof-where}).

\paragraph{The localisation heuristic.}
$\Delta V_N$ weights $\mathrm{d}Q$ by the $\mathrm{Beta}(N{+}1,2)$ kernel,
whose mode is $N/(N+1)$ (C4).  A feature at quantile $u_\star$ therefore
receives maximal weight when $N/(N+1)=u_\star$, i.e.\ at
$N_{\mathrm{loc}}=u_\star/(1-u_\star)$.

\paragraph{Proof of Lemma~\ref{lem:invisible}, and tightness.}
By Equation~\eqref{eq:bon-quantile},
$V_N-\tilde V_N=\E\big[(Q-\tilde Q)(U_{(N)})\big]
=\E\big[(Q-\tilde Q)(U_{(N)})\,\mathbf 1\{U_{(N)}>1-t\}\big]$, because the
integrand vanishes on $[0,1-t]$.  Hence
$|V_N-\tilde V_N|\le\|Q-\tilde Q\|_\infty\Pr(U_{(N)}>1-t)
=\|Q-\tilde Q\|_\infty\big(1-(1-t)^N\big)$, and $1-(1-t)^N\le Nt$ by
Bernoulli's inequality.  The bound is attained: if $Q-\tilde Q\equiv c$ on
$(1-t,1]$, then $V_N-\tilde V_N=c\,\big(1-(1-t)^N\big)\ge c\,(1-e^{-Nt})$.
A tail feature of height $c$ on the top-$t$ region is therefore invisible
for $Nt\ll1$ and fully expressed for $Nt\gg1$; the switch happens at
$N\approx1/t$, consistent with $N_{\mathrm{loc}}$ for $u_\star=1-t$.

\paragraph{Worked example: Student-$t$ errors.}
Let $S=R+\sigma T_\nu$.  Then
$\psi_\eps(t)=(\nu+1)t/(\nu\sigma^2+t^2)$, which is increasing on
$[-\sigma\sqrt\nu,\sigma\sqrt\nu]$ and decreasing on
$[\sigma\sqrt\nu,\infty)$.  By the localisation in
Appendix~\ref{app:proof-cov}, $G$ is nondecreasing whenever the window
$[s-r_{\max},s-r_{\min}]$ lies in $[-\sigma\sqrt\nu,\sigma\sqrt\nu]$, i.e.\
for $s\in[r_{\max}-\sigma\sqrt\nu,\,r_{\min}+\sigma\sqrt\nu]$, and
nonincreasing for $s\ge r_{\max}+\sigma\sqrt\nu$.  If
$r_{\max}-r_{\min}\le2\sigma\sqrt\nu$, the upper turning point of $G$
therefore lies in the band $[r_{\min}+\sigma\sqrt\nu,\,r_{\max}+\sigma\sqrt\nu]$.
When the reward range is small compared with $\sigma\sqrt\nu$, the score
exceedance probability there is
$\bar F_S\approx\Pr(T_\nu>\sqrt\nu)$, and the localisation heuristic
gives
\begin{equation}
\label{eq:student-localization}
 N_{\mathrm{loc}}\approx\frac{1}{\Pr(T_\nu>\sqrt\nu)}\approx4,\ 11,\ 26
 \qquad(\nu=1,3,5),
\end{equation}
independently of $\sigma$ to first order.  An exact evaluation of
Equation~\eqref{eq:bon-quantile} for $R\in\{0,1\}$, $\mu=1/2$ and
$\sigma\in\{1,3\}$ gives optimal budgets $\argmax_NV_N=5$, $11$ and
$22$--$24$ respectively.  The tail index places the peak; the noise scale
mainly sets its height.  As $\nu\to\infty$, $\Pr(T_\nu>\sqrt\nu)\to0$ and
the peak recedes to infinity, which recovers the monotone Gaussian case.

\paragraph{Proof of Proposition~\ref{prop:contam}.}
(a) The contamination label is independent of $R$ and of both errors, so
$S-R$ has the mixture density $p_\eps=(1-\pi)p_{\eps_0}+\pi p_{\eps_1}$
independently of $R$, and the model is additive.  Its hazard slope is
\begin{equation}
\label{eq:contamination-hazard}
 \psi_\eps=\frac{(1-\pi)(-p'_{\eps_0})+\pi\,\psi_{\eps_1}p_{\eps_1}}
               {(1-\pi)p_{\eps_0}+\pi\,p_{\eps_1}} .
\end{equation}
After dividing numerator and denominator by $p_{\eps_1}$, the numerator is
$(1-\pi)\,o(1)+\pi\,\psi_{\eps_1}\to0$ by the assumptions and the
denominator is at least $\pi$, so $\psi_\eps\to0$.
Theorem~\ref{thm:hazard} with $\lambda_\star=0$, whose Step~1 needs no
support assumption, gives $Q_\pi(u)\to m(0)=\mu$.  The mechanism is that
the heavy component, however rare, owns the extreme scores: the posterior
probability that a response at score $s$ is contaminated tends to one.

(b) Build both models on one probability space.  Let
$(R_i,\eps_{0,i},\eps_{1,i},B_i)_{i\le N}$ be independent across $i$, with
$B_i\sim\mathrm{Bernoulli}(\pi)$ the contamination flags.  Set
$S_i^0=R_i+\eps_{0,i}$ and $S_i^\pi=R_i+\eps_{0,i}$ if $B_i=0$, and
$S_i^\pi=R_i+\eps_{1,i}$ if $B_i=1$.  Each score vector has the correct law.
On $\{B_1=\dots=B_N=0\}$, which has probability $(1-\pi)^N$, the score
vectors coincide, and so do the selected indices (ties have probability
zero) and the selected rewards.  Otherwise the two selected rewards differ
by at most $r_{\max}-r_{\min}$.  Hence
$|V_N^\pi-V_N^0|\le(r_{\max}-r_{\min})\big(1-(1-\pi)^N\big)\le(r_{\max}-r_{\min})N\pi$.
\hfill$\square$

Part~(b) uses no property of $\eps_1$ whatsoever: rare mis-scoring is
invisible at small budgets whatever its form.  Part~(a) uses only the tail
of $\eps_1$ and is insensitive to $\pi$.  The rate $\pi$ therefore plays
exactly the role of $t$ in Lemma~\ref{lem:invisible}.

\paragraph{Proof of Corollary~\ref{cor:nocert}.}
For $N\le K$ and $\pi\le\delta/((r_{\max}-r_{\min})K)$, part~(b) gives
$|V_N^\pi-V_N^0|\le(r_{\max}-r_{\min})N\pi\le\delta$.  By part~(a),
$Q_\pi(u)\to\mu$, and the argument of Appendix~\ref{app:proof-return}
gives $V_N^\pi=\E[Q_\pi(U_{(N)})]\to\mu$.  \hfill$\square$

If $\eps_0$ is Gaussian and $r_{\max}$ is in the support, then
$V_N^0\to r_{\max}$ while $V_N^\pi\to\mu$.  The two curves are
indistinguishable to within $\delta$ on $N\le K$, yet their eventual gap is
$r_{\max}-\mu$, the entire value of selection.  The window of guaranteed
invisibility, $N\le\delta/((r_{\max}-r_{\min})\pi)$, is linear in $1/\pi$,
which is the precise sense in which halving the contamination rate doubles
it.

\section{Detailed Experimental Protocol}
\label{app:protocol}

\paragraph{Synthetic phase maps.}
The primary prior is bounded
$R\sim\mathrm{Uniform}[-\sqrt3,\sqrt3]$ ($\mu=0$, $r_{\max}=\sqrt3$), as
Theorem~\ref{thm:hazard} assumes; Bernoulli$(.5)$ (bounded, discrete) and
$\mathcal N(0,1)$ (unbounded, outside Theorem~\ref{thm:hazard}) are stress
tests.  For each prior, six target correlations
$\rho\in\{.5,.6,.7,.8,.9,.95\}$ are crossed with Gaussian,
variance-standardized Laplace, $t_\nu$ for $\nu\in\{30,10,6,4,3,2.5\}$, and
multiplicative $\eps=aRZ$ error.  Gaussian, Laplace, and $t_\nu$ are the
diverging-, constant-, and vanishing-influence regimes of
Theorem~\ref{thm:hazard}; the multiplicative error is the model
$S=R(1+aZ)$ of Section~\ref{sec:hetero}.  Noise scale $a$ matches the
pooled target correlation, so $\rho$ changes only where a tail becomes
visible.  Each of 162 cells uses 20,000 separately drawn $K=512$
candidate sets.  An interior peak is a downturn only when its selected
peak-to-512 contrast has a 95\% simultaneous max-$t$ interval above zero
over the tested grid.  Full grids and phase definitions follow below.  A separate 12-cell bounded-prior experiment
uses conditionally Gaussian errors with $\sigma_\beta(r)=\exp(\beta
r/\sqrt3)$ for $\beta\in\{-1,+1\}$ at the same six correlations; these
satisfy the assumptions of Proposition~\ref{prop:scale}, with unique scale
maximiser $r_\sigma=r_{\max}$ for $\beta=+1$ and $r_\sigma=r_{\min}$ for
$\beta=-1$.

\paragraph{Controlled real-model study.}
\label{sec:controlled-design}
Qwen2.5-7B-Instruct \citep{Qwen2024} generated 256 answers for each of 500
MATH-500 problems \citep{Hendrycks2021} and 1,319 GSM8K test questions
\citep{Cobbe2021}.  Four open reward models score the same fixed candidates:
Skywork-Reward-V2 Llama-3.2 1B/3B and Qwen3-4B
\citep{Liu2025Skywork}, plus Tulu-3 8B \citep{Lambert2024Tulu}.  Because
the candidates, and hence $\mu$, are held fixed, by
Equation~\eqref{eq:bon-quantile} every difference between conditions is a
difference in the calibration curve $Q$.  Besides QA and answer-only, MATH
controls for Tulu and Skywork Qwen3-4B preserve the user--assistant template
while either replacing the question with a fixed mask or deterministically
shuffling questions across answers.  Section~\ref{sec:theory} says what a
change in $Q$ implies, not what causes it; these controls separate missing
information from role/template shift as candidate causes.

\paragraph{External PPE validation.}
\label{sec:ppe-design}
PPE \citep{Frick2024} contributes 32 responses per prompt from four base
policies on MATH, GPQA, MMLU-Pro, MBPP+, and IFEval.  Thirteen learned RMs on
each task give 65 cells; NaiveVerbosityModel is retained only as one explicit
stress cell per task.  We exclude 31 nonnumeric Nemotron rows and record the
PPE revision, source hashes, exact exclusions, and policy strata.

\paragraph{Estimation and uncertainty.}
If $c_{(j)}$ is correctness at increasing score rank $j$ among $K$ candidates,
the exact finite-pool BoN expectation is
\begin{equation}
\label{eq:finite-pool-estimator}
 \widehat f(N)=\sum_{j=N}^{K}
 \frac{\binom{j-1}{N-1}}{\binom{K}{N}}c_{(j)},
\end{equation}
which is unbiased for $V_N$ of Proposition~\ref{prop:quantile} for every
$N\le K$; its increments have the nonnegative-kernel form of
Proposition~\ref{prop:finitepool}, so the rank profile both estimates $Q$
and witnesses any turn.  Ties are block averaged, which is exactly the
expectation under the uniform random tie breaking of
Section~\ref{sec:theory}; we also test candidate-index, reverse-index, and
repeated random tie orders.  Prompt bootstrap intervals preserve all
candidates; 2,000-replicate max-$t$ intervals control selected interior
contrasts within a cell.  Exact $K=256$ rank profiles determine controlled
$N^\star$.  PPE uses top-12.5\% excess $E_{\rm tail}$ (top four minus prompt
base correctness), the empirical mean of $E(u)$ over the top eighth of
scores, i.e.\ the band that budgets $N\approx8$ read by
Lemma~\ref{lem:invisible}; for binary correctness its sign is the sign of
$\log\Lambda$ averaged over that band (Proposition~\ref{prop:binaryLR}).

For the PPE intervention, two base-policy-stratified folds alternate as
calibration and test.  A negative calibration $E_{\rm tail}$ triggers the
calibration-optimal cap in $\{1,2,4,8,12,16,24,32\}$, a finite-pool version
of the stopping rule of Theorem~\ref{thm:shape}(b); otherwise $N=32$ is
retained.  One-sided cell tests are BH corrected over all 70 cells.
Diagnostic AUC, average precision, and Spearman correlation compare
$-E_{\rm tail}$, low rank accuracy, and score-tail spread; confidence
intervals resample cells while retaining both folds.  As a same-pool hedging
baseline, rank-Soft-BoN uses all 32 candidates, maps scores to tied
empirical quantiles (and so, like BoN, is invariant to monotone
recalibration, Proposition~\ref{prop:quantile}), and chooses its inverse
temperature on the calibration fold before held-out evaluation.

\paragraph{Synthetic grid.}
The complete budget grid is
\begin{equation}
\label{eq:synthetic-budget-grid}
\{1,2,3,4,6,8,12,16,24,32,48,64\}
\cup\{96,128,192,256,384,512\}.
\end{equation}
Student errors are variance standardized and the empirical second-moment
equation is solved for noise scale $a$ in every cell.  The selected-peak
interval uses 1,000 Gaussian-multiplier max-$t$ draws with seed 0.  A
non-downturn cell is called monotone when its empirical peak is $N=512$ and
plateau/boundary otherwise; these labels describe the observed window only.
Binary multiplicative scores have an atom at zero, making that row an explicit
tie/continuity stress test rather than an instantiation of
Proposition~\ref{prop:quantile}.
The 12 heteroskedastic cells use the same bounded Uniform prior, budget grid,
20,000 candidate sets, six target correlations, and 1,000-draw simultaneous
procedure.  Their model is
$S=R+a\exp(\beta R/\sqrt3)Z$ for $\beta\in\{-1,+1\}$; $a$ is solved per cell,
and seed 314159 deterministically seeds the cell sequence.

\paragraph{Controlled generation and scoring.}
Generation uses temperature $0.8$, top-$p=.95$, a 1,024-token output limit,
a 4,096-token context limit, and fixed per-prompt seeds.  Correctness follows
the repository's boxed-answer extraction and exact matching.  RMs use their
released chat templates, bfloat16 inference, and maximum length 4,096.  The
answer-only condition packages the answer as the sole user message, changing
both information and role relative to QA.  The
masked prompt is exactly ``[Question omitted]''.  Shuffling uses a fixed
nonzero cyclic offset determined by seed 42, ensuring that no answer receives
its original question.
For Tulu, visual $N^\star$ estimates over 16, 32, 50, 64, and 100 histogram
bins and local-linear bandwidths $.02,.05,.10$ are 3--4 for answer-only,
6--45 for QA, 6--59 when masked, and 5--25 when shuffled; the exact tied-rank
values are 3, 68, 82, and 18, respectively.

\paragraph{PPE provenance.}
The four base policies are Meta-Llama-3-8B-Instruct, Claude-3-Haiku-20240307,
Gemma-2-9B-IT, and GPT-4o-mini-2024-07-18.  Nonnumeric Nemotron exclusions are
23 MATH, 4 GPQA, 2 MMLU-Pro, 2 IFEval, and 0 MBPP+.  Every compact cell stores
its source-relative path and SHA-256 hash.  Source files contain 512 prompts
per task except 507 for MBPP+ before RM-specific exclusions.  Within-prompt tied score blocks
receive mean correctness before exact finite-pool integration.

\paragraph{Tail and cross-fit definitions.}
For PPE with $K=32$ and $P$ prompts,
\begin{equation}
\label{eq:ppe-tail-excess}
 E_{\mathrm{tail}}=\frac1P\sum_{i=1}^P\left(
 \frac14\sum_{j=K-3}^K c_{i,(j)}-
 \frac1K\sum_{j=1}^K c_{i,(j)}\right).
\end{equation}
The two folds are stratified within base policy.  A calibration-selected cap
is evaluated only on the opposite fold.  AUC labels a fold positive when the
held-out gain at its calibration-selected cap has a normal 95\% lower bound
above zero.  Average precision integrates precision over the distinct risk
thresholds and is compared with the empirical prevalence baseline.  Diagnostic
intervals use 2,000 cell-clustered bootstrap draws;
cross-fitted policy $p$-values are one-sided and BH corrected across all 70
cells.  Width sensitivity freezes these folds and uses $m\in\{2,4,8\}$;
leave-one-cell-out selection maximizes Spearman correlation with continuous
held-out gain in the other cells.  Tie sensitivity uses candidate-index,
reverse-index, and 200 uniformly randomized boundary-tie orders.

For rank-Soft-BoN, tied empirical quantiles $u_{ij}\in[0,1]$ give candidate
probabilities
\begin{equation}
\label{eq:rank-soft-bon}
 p_{ij}(\lambda)=\frac{\exp(\lambda u_{ij})}
 {\sum_{k=1}^{32}\exp(\lambda u_{ik})}.
\end{equation}
The calibration fold selects the smallest maximizing
$\lambda\in\{0,.125,.25,.5,1,2,4,8,16,32,64\}$ and the opposite fold reports
its exact expected correctness.  This quantile normalization follows the
score-rank step of HedgeTune but uses direct calibration utility rather than
its residual root finder.

\subsection{Correlation and finite-budget localization}
\label{app:synthetic-location}

\emph{Correlation sets location, not regime.}  For Uniform $t_{2.5}$,
the detected peak shifts from $N=12$ at $\rho=.5$ to $N=32$ at $\rho=.95$;
the corresponding drop contracts from $0.495$ [0.471, 0.520] to $0.296$
[0.282, 0.310].  For $t_6$, the peak moves to $N=128$ and the drop falls to
$0.0177$ [0.0118, 0.0235] at $\rho=.95$.  A smaller noise scale pushes the
quantiles where the heavy tail dominates further up, so larger budgets are
needed to read them (Lemma~\ref{lem:invisible}), while $\lambda_\star=0$
and hence the eventual return (Theorem~\ref{thm:hazard}) are unchanged.
Better global quality thus delays and attenuates hacking without changing
the error family.

\subsection{Heteroskedastic scale results}
\label{app:hetero-results}

\emph{Heteroskedastic scale.}  Proposition~\ref{prop:scale} predicts
$V_N\to\sqrt3$ for $\beta=+1$ and $V_N\to-\sqrt3<\mu=0$ for $\beta=-1$.  With
$\beta=+1$, all six cells peak empirically at $N=512$; with $\beta=-1$, four
have supported downturns.  For $\rho=.5$--$.8$, peaks move through
$N=4,6,12,32$ and drops range from $1.939$ to $0.600$, with simultaneous
lower bounds at least $0.573$.  Since no value exceeds
$r_{\max}=\sqrt3<1.939$, the largest drop already leaves BoN below random
selection, on the way to the predicted limit $r_{\min}$.  At $\rho=.9,.95$
the failure lies beyond the window.  Rank accuracy is still $0.828$ at
$\rho=.8$: high bulk quality coexists with a regime that ends below random,
which is the non-identifiability behind Corollary~\ref{cor:nocert}.

\subsection{Prior and multiplicative stress tests}
\label{app:stress-results}
\emph{Prior stress tests.}  Binary rewards remain within
Theorem~\ref{thm:hazard} through~\eqref{eq:binary-tilt} and yield 28
downturn, 19 plateau, and 7 monotone cells.  The unbounded Gaussian prior
lies outside Theorem~\ref{thm:hazard}, because the prior's own tail
competes with the error's, and yields 17/6/31.  Proposition~\ref{prop:cov}
still holds for both priors: Gaussian rows remain non-downturn, as does
$t_{30}$, and $t_{2.5}$ is the most robust downturn row.  Beyond the
bounded case, phase depends jointly on reward prior, error tail, and budget.

\emph{Multiplicative control.}  The multiplicative control is
monotone for Uniform and Gaussian priors and plateaus for binary rewards.
Its scale $\sigma(r)\propto|r|$ is maximised at \emph{both} edges of a
symmetric prior, so the uniqueness condition of
Proposition~\ref{prop:scale} fails, and the example $S=R(1+\eps)$ of
Section~\ref{sec:hetero} sits exactly at its boundary
$|r_{\min}|=r_{\max}$.  With the $O(s^2)$ scale terms tied, the $O(s)$
shift term breaks the tie in favour of $r_{\max}$, so extreme positive
scores still favour the good edge.  A scale that grows with $|R|$ is
therefore insufficient; what matters is which reward level dominates the
conditional upper tail.

\section{PPE Sensitivity Details}
\label{app:ppe-sensitivity}

At fixed folds, top-$m$ tail risk has AUC/AP $0.972/0.749$, $0.983/0.825$, and
$0.976/0.771$ for $m=2,4,8$.  By Lemma~\ref{lem:invisible}, these bands (the
top 1/16, 1/8, and 1/4 of scores) are those read at budgets of roughly 16, 8,
and 4.  All six negative signs survive two deterministic and 200 random tie
orders.  Nested selection favors $m=2$ without improving AP, supporting a
robust range, not an optimal width; tune $m\ge2$ on disjoint data.  With
$K=32$ candidates per prompt we do not estimate $\lambda_\star$ from
log-survival plots on PPE; $E_{\rm tail}$ targets $\Lambda$ at the top of the
score range directly (Proposition~\ref{prop:binaryLR}).  Best-of-Tails needs
endpoint-normalized rewards and ITP reference-policy likelihoods, absent from
these signed fixed-pool logits \citep{Hsu2026,Huang2025}; rank-Soft-BoN is
the feasible same-pool comparator.  The median learned cell has any tied score in 73.2\% of prompts and a
top-four-boundary tie in 7.0\%.  All six primary negative MBPP+ cells remain
negative for $m=2$; one additional GPQA cell crosses below zero.  For $m=8$,
five remain negative and Eurus moves to $+0.003$.  Leave-one-cell-out
continuous-gain selection chooses $m=2$ for 126/130 folds and $m=8$ for four,
with AP $0.749$.

\subsection{Relation to bulk ranking metrics}
\label{app:diagnostic-interpretation}

Rank accuracy is also strong (AUC/AP $0.972/0.747$); paired tail-minus-rank
intervals include zero for AUC and AP.  This does not contradict
Section~\ref{sec:theory}, whose claim is non-identifiability, not
non-correlation.  By the argument of Lemma~\ref{lem:invisible}, curves that
agree below quantile $1-\delta$ have bulk metrics within $O(\delta)$ of one
another yet may end in different regimes, and the synthetic $\beta=-1$,
$\rho=.8$ cell (rank accuracy $0.828$) is such a case.  On PPE, however, the
bad-tail cells are also low-rank-accuracy cells.  The broad AP intervals
expose the rarity and reject any claim that bulk ranking quality cannot
predict the problem in practice.

\section{Complete Controlled Results}
\label{app:controlled-results}

Table~\ref{tab:format-controls} reports the two structure-aware controls
alongside QA and answer-only.  Comparing masked-question with answer-only
separates question content from the chat-role shift; shuffled questions retain
the template while supplying deliberately mismatched semantics.

\subsection{Budget localization}
\label{app:controlled-localization}
Tulu's exact tied-rank $N^\star$ is 68 for QA, 3 for answer-only, 82 when
masked, and 18 when shuffled.  Because the kernel $u^N(1-u)$ of
Equation~\eqref{eq:marginal-value} peaks at $u=N/(N+1)$, these place the
turn of $Q$ near the 98.6th, 75th, 98.8th, and 94.7th score percentiles: the
answer-only failure sits deep in the bulk, whereas the other conditions
turn, if at all, only at the extreme top (Section~\ref{sec:where}).
Answer-only visual estimates stay in $\{3,4\}$ across five histogram grids and
three bandwidths; unsupported conditions vary more
(Appendix~\ref{app:protocol}).  Top-1\% excess is $+.0538,-.0310,+.0495,+.0231$
in the same order.

\begin{table}[htbp]
\caption{All original controlled conditions.  Drop is the selected interior
peak minus $N=256$ with a within-condition simultaneous max-$t$ interval.
\NA\ denotes a boundary maximum at $N=256$, with no interior peak contrast.}
\label{tab:controlled-full}
\centering
\footnotesize
\setlength{\tabcolsep}{4pt}
\begin{tabular}{lllrrr}
\toprule
Task & Input & RM & Rank acc. & Peak $N$ & Drop [95\% CI] \\
\midrule
\multirow{8}{*}{MATH-500}
 & QA & Skywork 1B & .746 & 24 & .0049 [-.0180,.0278] \\
 & QA & Skywork 3B & .772 & 160 & .0014 [-.0087,.0115] \\
 & QA & Skywork Qwen3-4B & .812 & 256 & \NA \\
 & QA & Tulu-3 8B & .741 & 64 & .0051 [-.0118,.0220] \\
 & answer-only & Skywork 1B & .712 & 24 & .0109 [-.0141,.0360] \\
 & answer-only & Skywork 3B & .735 & 256 & \NA \\
 & answer-only & Skywork Qwen3-4B & .799 & 256 & \NA \\
 & answer-only & Tulu-3 8B & .581 & 4 & .0597 [ .0316,.0878] \\
\midrule
\multirow{8}{*}{GSM8K}
 & QA & Skywork 1B & .706 & 12 & .0070 [-.0026,.0167] \\
 & QA & Skywork 3B & .731 & 12 & .0050 [-.0038,.0138] \\
 & QA & Skywork Qwen3-4B & .803 & 24 & .0028 [-.0044,.0100] \\
 & QA & Tulu-3 8B & .726 & 16 & .0031 [-.0064,.0126] \\
 & answer-only & Skywork 1B & .605 & 8 & .0046 [-.0053,.0145] \\
 & answer-only & Skywork 3B & .622 & 12 & .0042 [-.0064,.0149] \\
 & answer-only & Skywork Qwen3-4B & .732 & 32 & .0005 [-.0076,.0087] \\
 & answer-only & Tulu-3 8B & .569 & 4 & .0016 [-.0105,.0136] \\
\bottomrule
\end{tabular}
\end{table}

\section{PPE Reward Models and Additional Results}
\label{app:ppe-results}
\label{app:ppe-secondary}

Base-policy strata show joint policy--RM effects: Eurus tail excess on MBPP+
is $+.130$ for Meta-Llama-3-8B responses but $-.127$ for Claude-3-Haiku, with
peak-to-32 drops $.011$ and $.178$.  The other five negative cells also reverse
sign across policies.  Section~\ref{sec:theory} applies to the aggregate
curve, which averages these strata, so the top of $Q$ depends on the policy
mix.  In a way analogous to Proposition~\ref{prop:contam}, though not an
instance of it, a minority stratum with a bad tail can decide the top
quantiles, which precludes a task-independent safe budget here.

\paragraph{What occupies the MBPP+ bad tail?}
In terms of Proposition~\ref{prop:binaryLR}, this asks what drives the upper
tail of $p_0$, the score density of incorrect responses.  Released responses
show polished rather than malformed top-tail errors.  They are about 13\%
longer, add three lines, and across all six RMs are more likely to parse,
define the target function, contain a docstring, and place prose outside the
fence; repeated examples/assertions are 13.3 points less common.  These
shared-pool correlates are descriptive, not causal.

The 13 learned PPE score functions are ArmoRM-Llama3-8B-v0.1, Athene-RM-70B,
Athene-RM-8B, Eurus-RM-7B, Llama-3-OffsetBias-RM-8B,
Skywork-Reward-Gemma-2-27B, Skywork-Reward-Llama-3.1-8B, Starling-RM-34B,
Starling-RM-7B-alpha, InternLM2 reward models at 1.8B, 7B, and 20B, and
Nemotron-4-340B-reward.  PPE's NaiveVerbosityModel is retained only as a
stress baseline.  It has negative tail excess on all five tasks and a
supported BH-corrected cap gain on MATH, MMLU-Pro, and MBPP+, confirming that
the pipeline detects a deliberately shallow score rule without conflating it
with learned-RM prevalence.

\begin{table}[htbp]
\caption{Learned PPE cells with negative primary tail excess.  ``D'' marks the
strict interior-downturn definition; $q$ is for the cross-fitted cap policy
over all 70 cells.}
\label{tab:ppe-negative}
\centering
\footnotesize
\setlength{\tabcolsep}{4pt}
\begin{tabular}{lrrrrrl}
\toprule
MBPP+ RM & Rank acc. & $E_{\rm tail}$ & Peak & Drop & Cap gain [95\% CI] & $q$ \\
\midrule
Eurus-7B & .537 & -.014 & 2 & .059 D & .059 [.024,.093] & .0047 \\
Skywork-Llama-8B & .514 & -.028 & 2 & .043 & .039 [.001,.078] & .192 \\
Starling-7B & .387 & -.157 & 1 & .165 & .165 [.128,.202] & $<10^{-4}$ \\
InternLM2-1.8B & .464 & -.039 & 1 & .045 & .045 [.002,.088] & .161 \\
InternLM2-7B & .452 & -.114 & 1 & .171 & .171 [.132,.213] & $<10^{-4}$ \\
Nemotron-340B & .477 & -.068 & 1 & .076 & .076 [.030,.119] & .0027 \\
\bottomrule
\end{tabular}
\end{table}

\begin{table}[htbp]
\caption{Held-out mitigation comparison against hard BoN-32.  Intervals
resample learned task--RM cells and keep both folds together.}
\label{tab:mitigation-comparator}
\centering
\small
\setlength{\tabcolsep}{3.5pt}
\begin{tabular}{lrrrr}
\toprule
Population & Cells & Tail cap $\Delta$ & Rank-Soft-BoN $\Delta$ & Cap $-$ soft \\
\midrule
All learned & 65 & .0085 [.0019,.0168] & .0068 [-.0003,.0156] & .0017 [-.0003,.0036] \\
Negative tail & 6 & .0924 [.0519,.1383] & .0916 [.0511,.1339] & .0009 [.0000,.0020] \\
\bottomrule
\end{tabular}
\end{table}

Among learned cells, the median prompt-level frequency of any tied score is
73.2\%, but the median frequency of a tie intersecting the top-four boundary
is only 7.0\%.  The six negative cells remain negative under both deterministic
orders and all 200 randomized orders.

\section{Descriptive Extreme-Answer Features}

For the controlled MATH pool, we report four predefined descriptive answer
features: log character count, number of equation markers, presence of a
\texttt{\\boxed} marker, and whether the final answer is extractable.  For
each score format, the analysis gives within-prompt Spearman correlation with
score and correctness contrasts in the top 1\%, 5\%, and 10\% of score ranks,
with prompt-bootstrap intervals.  Table~\ref{tab:features} summarizes the
Tulu top-1\% comparison.  In the failing answer-only condition, extreme
answers are shorter, contain fewer equation markers (contrast $-0.88$
[$-1.56,-0.23$]), and are less likely to contain boxed or extractable final
answers.  Masking reverses the correctness and final-answer-marker contrasts.
The failure is therefore not a simple verbosity effect.  These associations
only describe which response forms occupy an extreme tail; they do not
identify a causal reward-model shortcut.

\begin{table}[htbp]
\caption{Descriptive Tulu MATH top-1\% features.  $\rho_L$ is mean
within-prompt Spearman correlation between score and log character count;
$\Delta_L$ and $\Delta_B$ compare top-tail versus remaining log length and
boxed-marker frequency; $E_1$ is correctness excess.}
\label{tab:features}
\centering
\small
\setlength{\tabcolsep}{4pt}
\begin{tabular}{lrrrr}
\toprule
Input & $\rho_L$ & $\Delta_L$ [95\% CI] & $\Delta_B$ [95\% CI] & $E_1$ [95\% CI] \\
\midrule
QA & -.106 & -.020 [-.031,-.009] & +.084 [.069,.100] & +.053 [.034,.073] \\
Answer-only & -.144 & -.029 [-.040,-.018] & -.026 [-.041,-.011] & -.031 [-.048,-.013] \\
Masked & -.042 & +.006 [-.005,.018] & +.067 [.052,.082] & +.048 [.029,.066] \\
Shuffled & +.050 & +.027 [.016,.038] & +.024 [.011,.037] & +.023 [.006,.038] \\
\bottomrule
\end{tabular}
\end{table}

For MBPP+, code is extracted from the longest Python fence (or the full
response if unfenced) and parsed with Python's AST.  The expected function name
comes from the released reference code.  Table~\ref{tab:mbpp-features} reports
RM-cell-level contrasts; no confidence interval treats the six repeated views
of the shared response pool as independent.

\begin{table}[htbp]
\caption{MBPP+ incorrect top-tail minus other incorrect response features over
the six negative-RM cells.  Binary-feature contrasts are proportions.}
\label{tab:mbpp-features}
\centering
\small
\begin{tabular}{lrr}
\toprule
Feature & Median contrast & RM-cell range \\
\midrule
Log character count & +.121 & [.025,.150] \\
Line count & +2.986 & [.424,3.850] \\
Syntax valid & +.018 & [.007,.022] \\
Defines target function & +.021 & [.000,.036] \\
Target-function docstring & +.029 & [.007,.037] \\
Examples or assertions & -.133 & [-.217,-.112] \\
Prose outside code fence & +.128 & [.105,.148] \\
Comment fraction & -.004 & [-.038,-.001] \\
\bottomrule
\end{tabular}
\end{table}

\end{document}